\documentclass{lmcs} 
\pdfoutput=1
\usepackage[utf8]{inputenc}

\usepackage{lastpage}
\lmcsdoi{22}{1}{19}
\lmcsheading{}{\pageref{LastPage}}{}{}%
{Feb.~14,~2025}{Mar.~06,~2026}{}

\keywords{Personalization, Federated Learning, Aggregate Computing, Field-based Coordination}

\usepackage{hyperref}
\usepackage[inline]{enumitem}
\usepackage{mytodonotes}
\usepackage{subcaption}

\usepackage[capitalise]{cleveref}
\usepackage{algorithm}
\usepackage{algpseudocode}
\usepackage{listings}
\usepackage{tcolorbox}
\usepackage{amsmath}
\usepackage{amsfonts}
\usepackage{acronym}
\usepackage{booktabs}
\usepackage{fontawesome5}
\usepackage{pifont}
\usepackage{mathtools}
\usepackage{amsthm}
\DeclareMathOperator*{\argmin}{arg\,min}
\newtheorem{assumption}{Assumption}
\lstdefinelanguage{scala}{
  keywords={abstract,case,catch,class,def,%
    do,else,extends,false,final,finally,%
    for,if,implicit,import,match,mixin,%
    new,null,object,override,package,%
    private,protected,requires,return,sealed,%
    super,this,throw,trait,true,try,lazy,%
    type,val,var,while,with,yield,forSome},
  otherkeywords={=>,<-,<\%,<:,>:,\#},
  sensitive=true,
  columns=fullflexible,
  morecomment=[l]{//},
  morecomment=[n]{/*}{*/},
  morestring=[b]",
  stringstyle=\ttfamily\color{red!50!brown},
  showstringspaces=false,
  morestring=[b]',
  morestring=[b]""",
  basicstyle=\sffamily\lst@ifdisplaystyle\footnotesize\fi\ttfamily,
  emphstyle=\sffamily\bfseries\ttfamily
}
\definecolor{ddarkgreen}{rgb}{0,0.5,0}
\lstdefinelanguage{scafi}{
  frame=single,
  basewidth=0.5em,
  language={scala},
  keywordstyle=\color{blue}\textbf,
  commentstyle=\color{ddarkgreen},
  keywordstyle=[2]\color{teal}\textbf,
  keywords=[2]{rep,nbr,foldhood,foldhoodPlus,aggregate,branch,spawn,mux,mid},
  keywordstyle=[3]\color{gray},
  keywords=[3]{Me,AroundMe,Everywhere,Forever}, 
  keywordstyle=[4]\color{red}\textbf,
  keywords=[4]{in,out,rd},
  keywordstyle=[5]\color{violet},
  keywords=[5]{evolve,when,andNext,workflow,G,C,S,broadcast,gossip},
  keywordstyle=[6]\color{orange},
  keywords=[6]{Available,Serving,Done,Waiting,Removing,None,Set}
}

\theoremstyle{plain} 

\makeatletter
\AtBeginDocument
 {
   \def\ltx@label#1{\cref@label{#1}}
   \def\label@in@display@noarg#1{\cref@old@label@in@display{#1}}
    \def\label@in@mmeasure@noarg#1{%
      \begingroup%
      \measuring@false%
      \cref@old@label@in@display{#1}
      \endgroup
    }%
 } %
\makeatother

\newlist{questions}{enumerate}{2}
\setlist[questions,1]{label=(RQ\arabic*),ref=RQ\arabic*}
\setlist[questions,2]{label=(\alph*),ref=\thequestionsi(\alph*)}

\def\eg{{\em e.g.,}}

\def\ie{{\em i.e.,}}

\acrodef{fl}[FL]{federated learning}
\acrodef{iid}[IID]{independent and identically distributed}
\acrodef{fbfl}[FBFL]{Field-Based Federated Learning}

\begin{document}

\title[Field-Based Coordination for Federated Learning]{FBFL: A Field-Based Coordination Approach for Data Heterogeneity in Federated Learning}

\author[D.~Domini]{Davide Domini\lmcsorcid{0009-0006-8337-8990}}[a]

\author[G.~Aguzzi]{Gianluca Aguzzi\lmcsorcid{0000-0002-1553-4561}}[a]

\author[L.~Esterle]{Lukas Esterle\lmcsorcid{0000-0002-0248-1552}}[b]

\author[M.~Viroli]{Mirko Viroli\lmcsorcid{0000-0003-2702-5702}}[a]

\address{ Alma Mater Studiorum—Università di Bologna, Via dell'Università, 50, Cesena (FC), Italy }	
\email{davide.domini@unibo.it, gianluca.aguzzi@unibo.it, mirko.viroli@unibo.it} 

\address{ Aarhus  University, Aarhus, Denmark }
\email{lukas.esterle@ece.au.dk} 




\begin{abstract}
In the last years, \Ac{fl} has become a popular solution to train machine learning models
 in domains with high privacy concerns.
%
%
However, 
 \ac{fl} scalability and performance face significant challenges 
 in real-world deployments where data across devices are non-independently and identically 
 distributed (non-\acs{iid}).
The heterogeneity in data distribution frequently arises from spatial distribution of devices,
 leading to degraded model performance in the absence of proper handling.
Additionally, 
 \ac{fl} typical reliance on centralized architectures introduces bottlenecks
 and single-point-of-failure risks, particularly problematic at scale or in dynamic environments.

To close this gap, 
we propose \Ac{fbfl}, a novel approach leveraging 
 macroprogramming and field coordination to address these limitations through:
\begin{enumerate*}[label=(\roman*)]
  \item distributed spatial-based leader election for personalization to mitigate non-\acs{iid} data challenges; and
  \item construction of a self-organizing, hierarchical architecture using advanced macroprogramming patterns. 
\end{enumerate*}
Moreover, \ac{fbfl} not only overcomes the aforementioned limitations, but also enables the development 
 of more specialized models tailored to the specific data distribution in each subregion.

This paper formalizes \ac{fbfl} and evaluates it extensively using 
 MNIST, FashionMNIST, and Extended MNIST datasets. 
We demonstrate that, when operating under \acs{iid} data conditions, \ac{fbfl} performs 
 comparably to the widely-used FedAvg algorithm.
Furthermore, in challenging non-\acs{iid} scenarios, \ac{fbfl} not only outperforms FedAvg but also 
 surpasses other state-of-the-art methods, namely FedProx and Scaffold, which have been 
 specifically designed to address non-\acs{iid} data distributions.
Additionally, we showcase the resilience of \ac{fbfl}'s self-organizing hierarchical architecture against server failures.
\end{abstract}

\maketitle

\section{Introduction}\label{sec:intro}
\paragraph{\emph{Research Context.}} 
Machine learning requires large datasets to train effective and accurate models. 
Typically, data are gathered from various sources into one central server where the training process occurs.
However, centralizing data on a single device poses significant privacy challenges.
These concerns arise not only from the need to share sensitive information but also from the 
 heightened risk of storing all data in a single location, which, if compromised, could result in 
 large-scale data exposure.
\Acf{fl} has emerged as a promising solution for training machine learning models 
 in scenarios where \emph{data privacy} is a primary concern,
 enabling devices to 
 collaboratively learn a shared model while keeping their data local~\cite{DBLP:conf/aistats/McMahanMRHA17}.
This paradigm not only alleviates the necessity for central data storage but also addresses 
 privacy and efficiency concerns inherent in traditional systems with centralized learning.

\paragraph{\emph{Research Gap.}}
Despite its advantages, in the current landscape of \ac{fl}, training is distributed, but model 
 construction is still predominantly centralized, posing challenges in scenarios characterized 
 by spatial dispersion of devices, 
 heightened risk of single points of failure, and naturally distributed datasets. 
Existing peer-to-peer solutions attempt to tackle these concerns; 
 however, they often overlook the spatial distribution of devices 
 and do not exploit the benefits of semantically similar knowledge 
 among nearby devices~\cite{DBLP:conf/sac/MalucelliDAV25}. 
This builds on the assumption that devices in spatial proximity have similar experiences and 
 make similar observations, 
 as the phenomena to capture are intrinsically context dependent~\cite{esterle2022deep}.
Moreover, existing approaches often lack the flexibility to seamlessly transition 
 between fully centralized and fully decentralized aggregation methods. 
This limitation highlights the potential role of advanced coordination 
 models and programming languages in bridging this gap.

\paragraph{\emph{Contribution}}
For these reasons, in this paper, 
 we introduce \acf{fbfl}, a novel approach that leverages computational fields (\ie{} global maps from devices to values) as key abstraction 
 to facilitate device coordination \cite{VBDACP-COORDINATION2018,DBLP:journals/corr/Lluch-LafuenteL16} in \ac{fl}. 
By field-based coordination, global-level system behavior can be captured declaratively, 
 with automatic translation into single-device local behavior.
 
We find that this approach
 offers a versatile and scalable solution to \ac{fl}, 
 supporting the formation of \emph{personalized model zones}---spatially-based regions where devices with similar data distributions collaboratively train specialized models tailored to their local context, rather than a single global model shared across all participants.
This represents a form of \emph{personalized federated learning}, 
 where instead of learning one universal model that may perform poorly on diverse local data distributions, the system creates multiple specialized models, 
 each optimized for the specific characteristics of data within its respective zone.
Most specifically, our approach actually relies on known field-based algorithms of information 
 diffusion and aggregation developed in the context of aggregate computing \cite{VABDP-TOMACS2018}, 
 ultimately defining what we can define ``fields of Machine Learning (sub)models''.
This method enables dynamic, efficient model aggregation without a centralized authority, 
 thereby addressing the limitations of current \ac{fl} frameworks.
Therefore, our contributions are twofold:
\begin{itemize}
  \item We demonstrate that field coordination enables performance comparable to centralized approaches under \ac{iid} data settings;
  \item We exploit advanced aggregate computing patterns to efficiently create a self-organizing hierarchical
   architecture that group devices based on spatial proximity, 
   improving performance under non-\ac{iid} data settings.
\end{itemize}
By ``self-organizing hierarchical'' we mean a hybrid architecture based on peer-to-peer interactions, but which elects leaders in a distributed manner. 
These leaders will act as aggregators of local models pertaining to sub-portions of the system.
We evaluate our approach in a simulated environment, 
 where well-known computer vision datasets--MNIST~\cite{lecun2010mnist}, FashionMNIST~\cite{DBLP:journals/corr/abs-1708-07747}, 
 and Extended MNIST~\cite{DBLP:journals/corr/CohenATS17}--are 
  synthetically split to create non-\ac{iid} partitions.
As baselines, we employ three state-of-the-art algorithms, namely:
 FedAvg~\cite{DBLP:conf/aistats/McMahanMRHA17}, 
 FedProx~\cite{DBLP:conf/mlsys/LiSZSTS20}, 
 and Scaffold~\cite{DBLP:conf/icml/KarimireddyKMRS20}.
%
Our findings indicate that this field-based strategy not only matches the performance of existing 
 methods but also provides enhanced scalability and flexibility in \ac{fl} implementations.

\paragraph{\emph{Paper Structure.}} 
This manuscript is an extended version of the conference paper~\cite{DBLP:conf/coordination/DominiAEV24}, providing:
\begin{enumerate*}[label=(\roman*)]
  \item a more extensive and detailed coverage of related work;
  \item an expanded formalization which adds the description of the self-organizing hierarchical
   architecture proposed by this paper;
  \item a largely extended experimental evaluation adding more datasets (\ie{} MNIST, Fashion
   MNIST and Extended MNIST) and more baselines (\ie{} FedAvg, FedProx and Scaffold); and
  \item a new experiment for testing the resilience of the 
  self-organizing hierarchical architecture simulating aggregators failures.
\end{enumerate*}
The remainder of this paper is organized as follows: 
 \Cref{sec:rqs} states the research questions. 
 \Cref{sec:background} reviews the necessary background on federated learning and field-based coordination. 
 \Cref{sec:related} presents related works, the baseline algorithms, and our motivation. 
 \Cref{sec:fbfl} introduces our Field-Based Federated Learning approach: we formalize the problem (\Cref{sec:problem-formulation}), describe the distributed algorithm (\Cref{sec:fbfl-algo}), and provide implementation details (\Cref{sec:learningprocess}). 
 \Cref{sec:evaluation} reports the experimental evaluation and discussion. 
 \Cref{sec:limitations} discusses limitations. 
 Finally, \Cref{sec:conclusions} concludes and outlines future work.

\section{Research Questions}\label{sec:rqs}

The complexity of modern systems where \acf{fl} can be applied, 
 such as the edge-cloud compute continuum, 
 is rapidly increasing. This 
 poses significant challenges in terms of scalability, adaptability, and contextual relevance~\cite{DBLP:journals/fgcs/PrigentCAC24}. 
Field-based approaches have demonstrated notable advantages in addressing such complexity in various domains, 
 offering robust solutions in both machine learning~\cite{DBLP:conf/icdcs/AguzziCV22,DBLP:journals/scp/DominiCAV24},
  software engineering~\cite{DBLP:conf/acsos/CortecchiaPCC24} contexts
  and real world use cases~\cite{DBLP:conf/coordination/AguzziBBCCDFPV25}. 
However, research on field-based methodologies within \ac{fl}, particularly in the area 
  of personalized \ac{fl}~\cite{DBLP:conf/acsos/DominiAFVE24,DBLP:conf/acsos/Domini24,DOMINI2026101841}, is still at an early stage.
In particular, 
current approaches still suffer from three key limitations: reliance on centralized aggregation (limiting scalability and resilience), 
difficulty handling highly non-\ac{iid} data, 
and lack of mechanisms to exploit spatial correlation in data distributions.
Our proposed \ac{fbfl} framework directly targets these limitations.
To assess its effectiveness, we focus on the following research questions:

\begin{questions}
  \item How does the Field-Based Federated Learning approach perform compared 
   to centralized \ac{fl} under \ac{iid} data? \label{itm:rq1}
  \item Can we increase accuracy by introducing personalized learning zones where \emph{learning devices} are grouped based 
   on similar experiences as often observed in spatially near locations? 
   More precisely, what impact does this have on heterogeneous 
   and non-\ac{iid} data distributions?\label{itm:rq2} 
   \item What is the effect of creating a self-organizing hierarchical architecture through 
   field coordination on federated learning in terms of 
   resilience, adaptability and fault-tolerance? \label{itm:rq3}
\end{questions}

\section{Background}\label{sec:background}

\subsection{Federated learning}\label{sec:background-fl}

\begin{figure*}
  \centering
  \begin{subfigure}{0.49\textwidth}
      \centering
      \includegraphics[width=\textwidth]{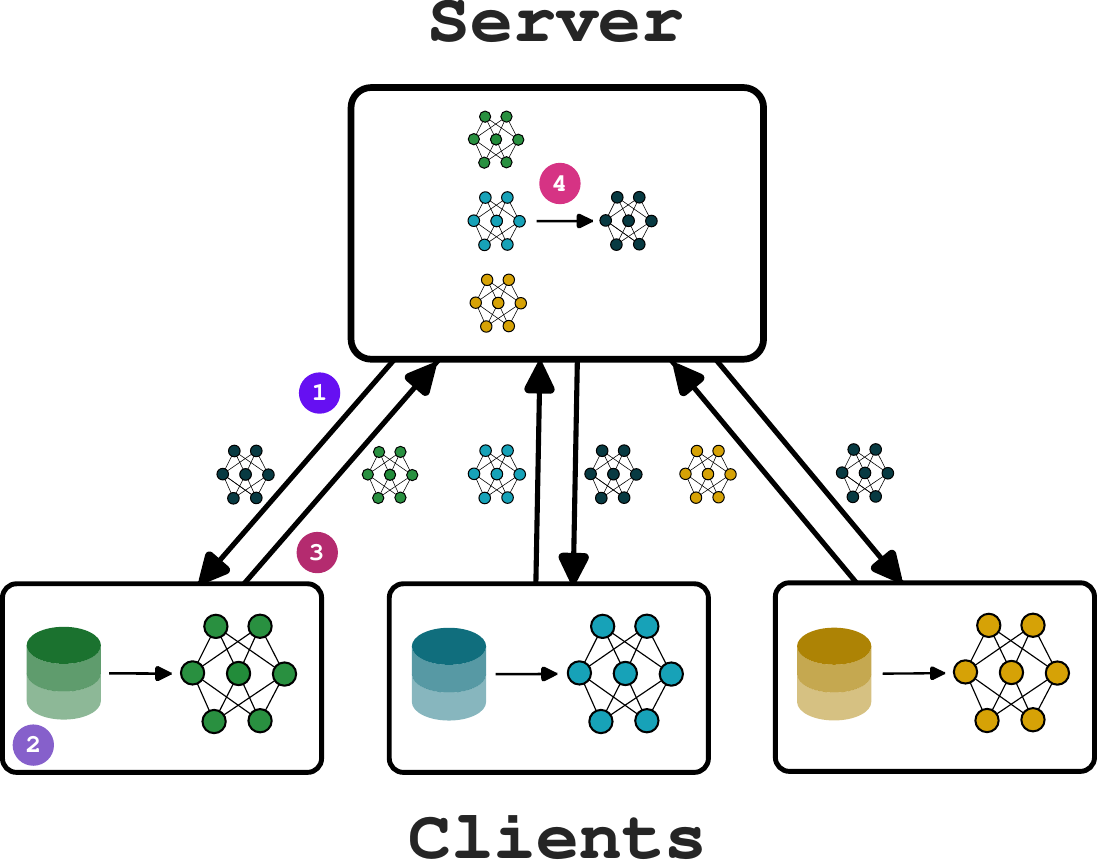}
      \caption{Centralized federated learning schema.}
      \label{fig:cfl}
  \end{subfigure}
  \begin{subfigure}{0.49\textwidth}
    \centering
    \includegraphics[width=\textwidth]{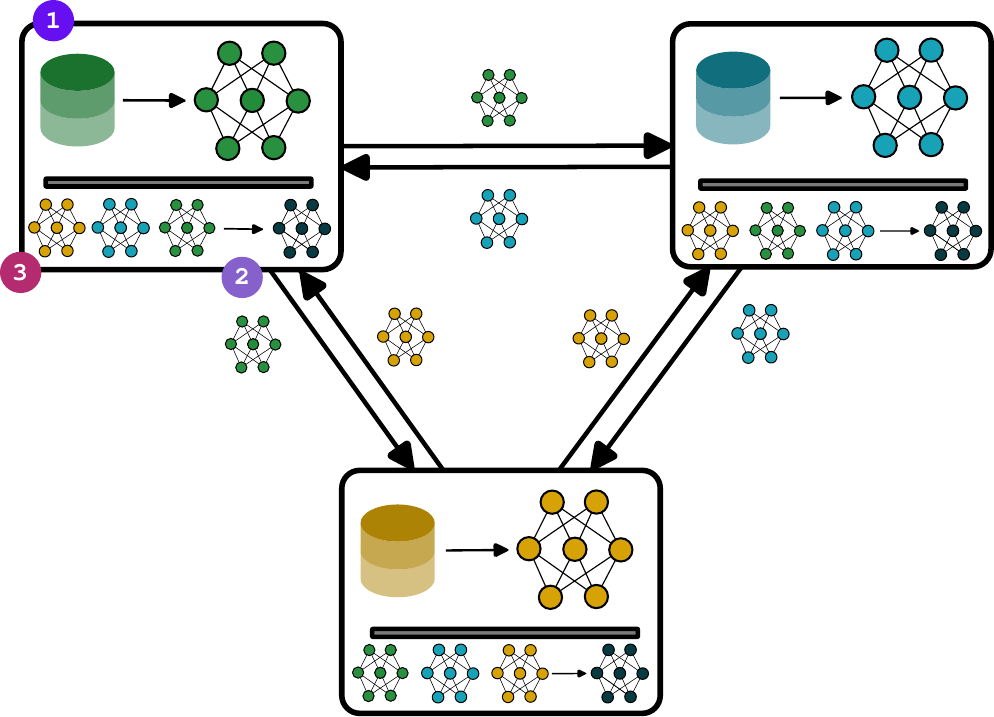}
    \caption{Peer-to-peer federated learning schema.}
    \label{fig:p2pfl}
  \end{subfigure}
  \caption{ On the left, a visual representation of centralized federated learning. 
  In the first phase, the server
  shares the centralized model with the clients. In the second
  phase, the clients perform a local learning phase using data that
  is not accessible to the server. In the third phase, these models
  are communicated back to the central server, and finally, in the
  last phase, there is an aggregation algorithm. 
  On the right, the P2P federated learning schema.
  Differently from the centralized version, there is no central server. 
  Each client sends its local model to all the other clients, then the aggregation
  is performed locally. 
  }
  \label{fig:tbd}
\end{figure*}

Federated learning (FL)~\cite{DBLP:conf/aistats/McMahanMRHA17} is a machine learning technique 
 introduced to collaboratively train a joint model using multiple, 
 potentially \emph{spatially distributed}, devices.
In general, in the context of FL, a ``model'' refers to any machine learning 
 model---such as decision trees, support vector machines, or neural networks---that can be 
 trained on data to perform increasingly accurate predictions. 
These are statistical models whose parameters are iteratively updated based on local 
 data to minimize prediction error (\ie{} the loss function). 
In this work, we focus specifically on neural networks as the model class of interest, 
 given their widespread adoption and strong empirical performance in a variety of learning tasks.

The federation typically consists of multiple \emph{client} devices and 
 one \emph{aggregation} server, which may be located in the cloud. 
The key idea behind \ac{fl} is that the data always remains on the device to which it belongs: 
 devices only share the weights of their models (\ie{} the parameters of the neural network), 
 thus making \ac{fl} an excellent solution in contexts where \emph{privacy} is a crucial aspect 
 (\eg{} in health-care applications~\cite{DBLP:journals/jhir/XuGSWBW21,DBLP:journals/csur/NguyenPPDSLDH23}) 
 or where data are \emph{naturally distributed}, and their volume makes it infeasible to transfer them 
 from each client device to a centralized server.
Federated learning can be performed in different ways~\cite{DBLP:journals/tist/YangLCT19}. 
In horizontal FL (a.k.a., sample-based), participants share the same input features but hold distinct sample sets.
They generally follow four steps (see \Cref{fig:cfl}):
\begin{enumerate}
  \item \emph{model distribution}: the server distributes the initial global model to each device. 
   This step ensures that all devices start with the same model parameters before local training begins;
  \item \emph{local training}: each device trains the received model on its local dataset for a specified number of epochs. 
   This step allows the model to learn from the local data while keeping the data on the device, thus preserving privacy;
  \item \emph{model sharing}: after local training, each device sends its updated model parameters back to the server. 
   This step involves communication overhead but is crucial for aggregating the learned knowledge from all devices;
  \item \emph{model aggregation}: the server collects all the updated model parameters from the devices and combines them using an aggregation algorithm (\eg{} averaging the weights). 
   This step produces a new global model that incorporates the knowledge learned by all devices. 
   The process then repeats for a predefined number of rounds or until convergence.
\end{enumerate}
Despite its advantages, \ac{fl} faces several challenges, 
such as \emph{non-\ac{iid}} data distribution and \emph{resilience} to server failures, discussed in the following section.

\subsubsection{Challenges}\label{sec:challenges}

\paragraph{\emph{Resilience}}
Exploring the network topology is crucial, 
 as the configuration of device communication can significantly influence \ac{fl} performance. 
Various structures, such as \emph{peer-to-peer} (P2P), \emph{hierarchical}, and \emph{centralized} networks, 
 offer diverse benefits and challenges. 
Traditionally, \ac{fl} systems have relied on a centralized server for model aggregation. 
However, this setup poses scalability challenges and risks introducing a single point of failure. 
In response, recent advancements (\eg{}~\cite{DBLP:journals/jpdc/HegedusDJ21,DBLP:conf/dsn/WinkN21,DBLP:conf/icc/Liu0SL20}) 
 have embraced P2P networks to address these limitations, 
 going towards what is called decentralized federated learning---see~\Cref{fig:p2pfl}.
P2P architectures, devoid of a central authority, enhance scalability and robustness through 
 the utilization of gossip~\cite{DBLP:conf/aistats/VanhaesebrouckB17} 
 or consensus algorithms~\cite{DBLP:journals/iotj/SavazziNR20} for model aggregation.

\paragraph{\emph{Data heterogeneity}}
Federated learning is particularly effective when the data distribution across 
 devices is \emph{independent and identically distributed} (\ac{iid})~\cite{DBLP:conf/middleware/NilssonSUGJ18}, 
 namely the users experience the same data distribution. 
For instance, in a text prediction application, all users may have similar writing styles.
However, in real-world scenarios, data distribution is often non-\ac{iid}, 
 where devices have different data distributions. 
The skewness of data can be categorized in various ways based on how they 
 are distributed among the clients~\cite{DBLP:conf/icde/LiDCH22,DBLP:journals/ftml/KairouzMABBBBCC21}. 
The main categories include:
\begin{enumerate*}[label=(\roman*)]    
    \item \emph{feature skew}: all clients have the same labels but different feature 
     distributions (\eg{} in a handwritten text classification task, we may have the same letters 
     written in different calligraphic styles);
    \item \emph{label skew}: each client has only a subset of the classes; and
    \item \emph{quantity skew}: each client has a significantly different amount of data compared to others.
\end{enumerate*}
In this work, we focus on label skew. 
More in detail, given the features $x$, a label $y$, and 
 the local distribution of a device $i$, denoted as:
\begin{equation}
  \mathbb{P}_i(x,y) = \mathbb{P}_i(x \mid y) \cdot \mathbb{P}_i(y)
\end{equation}
the concept of label skew, between two distinct devices $k$ and $j$, is defined as follows~\cite{DBLP:journals/pami/HuangYSWLDY24}:
\begin{equation}
  \mathbb{P}_k(y) \neq \mathbb{P}_j(y) \text{ and } \mathbb{P}_k(x \mid y) = \mathbb{P}_j(x \mid y)
\end{equation}
In other words, different clients have distinct label distributions (\ie{} $\mathbb{P}_i(y)$) 
 while maintaining the same underlying feature distribution for each label (\ie{} $\mathbb{P}_i(x\mid y)$).

This data heterogeneity can lead to \emph{model drift}~\cite{DBLP:conf/nips/WangLLJP20} during training,
 where the model's performance degrades as training progresses, 
 potentially causing slow or unstable convergence.
In particular, this phenomenon occurs when clients with significantly different 
 data distributions produce highly divergent model updates, 
 each optimized for their own local objectives. 
As a result, the aggregated global model may exhibit large parameter shifts across rounds, 
 which can harm overall performance and hinder convergence.
To address these challenges, several approaches have been proposed, 
 such as \emph{FedProx}~\cite{DBLP:conf/mlsys/LiSZSTS20} and \emph{Scaffold}~\cite{DBLP:conf/icml/KarimireddyKMRS20}.

\subsection{Field-Based Coordination}\label{sec:background-fbc}

\emph{Coordination based on fields}~\cite{DBLP:conf/forte/AudritoVDPB19} (or \emph{field-based coordination}) employs 
 a methodology where computations are facilitated through the concept of \emph{computational fields} (\emph{fields} in brief), 
 defined as distributed data structures that evolve over time and map locations with specific values.
This method draws inspiration from foundational works such as Warren's \emph{artificial potential fields} 
(namely, a technique where robots navigate by following gradient-based force fields that attract them to goals while repelling them from obstacles)~\cite{DBLP:conf/icra/Warren89}
 and the concept of \emph{co-fields} by Zambonelli et al.~\cite{DBLP:journals/pervasive/MameiZL04}.
Specifically, in the context of co-fields, 
 these computational fields encapsulate contextual data, 
 which is sensed locally by agents and disseminated by either the agents themselves or the infrastructure following a specific distribution rule.

In our discussion, \emph{coordination based on fields} refers to a distinct macroprogramming and computation framework, 
 often referred to as \emph{aggregate computing} (AC)~\cite{VBDACP-COORDINATION2018}.
This framework enables the programming of collective, 
 self-organizing behaviors through the integration of functions that operate on fields, 
 assigning computational values to a collection of individual agents 
 (as opposed to locations within an environment).
Thus, fields facilitate the association 
 of a particular domain of agents with their sensory data,
 processed information, and instructions for action within their environment.
Computed locally by agents yet under a unified perspective, 
 fields enable the representation of collective instructions (\eg{} a velocity vector field or a field of machine learning model parameters).
To comprehend field-based computing fully, 
 we highlight the system model and the programming model in the following sections.

\subsubsection{System Model.}\label{ssec:background:sysmodel}
An aggregate system may be defined as a collection of \emph{agents} (or \emph{nodes}) 
 that interact within a shared environment to achieve a common goal.

To better understand the system's behavior, 
 we must first define the system's structure (\ie{} the agents and their relationships), 
 the agents' interaction (\ie{} how they communicate),
 and their behavior within the environment (\ie{} how they process information and act).

\paragraph{\emph{Structure.}}
An \emph{agent} represents an autonomous unit furnished with \emph{sensors} and \emph{actuators} to interface 
 with either a logical or physical \emph{environment}.
From a conceptual standpoint, an agent possesses \emph{state}, capabilities for \emph{communication} with 
 fellow agents, and the ability to \emph{execute} simple programs.
An agent's \emph{neighborhood} is composed of other \emph{neighbor} agents, forming a connected network that can be 
 also represented as a graph---see \Cref{sec:problem-formulation} for more details.
The composition of this network is determined by a \emph{neighboring relationship}, designed based on the specific application 
 needs and constrained by the physical network's limitations.
Commonly, neighborhoods are defined by physical connectivity or spatial proximity, allowing for communication directly or 
 via infrastructural support, based on certain proximity thresholds.

\paragraph{\emph{Interaction.}}
Agents interact by asynchronously sending messages to neighbors,
 which can also occur stigmergically---through indirect coordination via environmental modifications---through sensors and actuators.
The nature and timing of these messages depend on the agent's programmed behavior.
Notably, our focus on modelling continuous, 
 self-organizing systems suggest frequent interactions relative to the dynamics of the problem and environment.

\paragraph{\emph{Behavior.}}
%
Based on the interaction of agents, an agent's behavior unfolds through iterative \emph{execution rounds}, 
 with each round encompassing the steps below, albeit with some flexibility in the actuation phase:
\begin{enumerate}
\item \emph{Context acquisition:} agents accumulate context by considering both their prior state and the
 latest inputs from sensors and neighboring messages.
\item \emph{Computation:} agents process the gathered context, resulting in 
 (i) an \emph{output} for potential actions, and 
 (ii) a \emph{coordination message} for neighborly collective coordination.
\item \emph{Actuation and communication:} agents execute the actions as specified by the output and 
 distribute the coordination message across the neighborhood.
\end{enumerate}
By cyclically executing these sense-compute-act rounds, 
 the system exhibits a self-organizing mechanism 
 that integrates and processes fresh data from both the environment and the agents, 
 typically achieving self-stabilization

\subsubsection{Programming model.}\label{ssec:background:progmodel}
This system model establishes a foundation for collective adaptive behavior,
 necessitating an in-depth elucidation of the ``local computation step'', 
 facilitated by a \emph{field-based programming model}. 
This model orchestrates the collective behavior through a \emph{field-based program},
 executed by each agent in adherence to the prescribed model. 
Field calculus defines the computation as a composition of \emph{function operations} on fields, and any variants of it allow the developers to express at least 
 i) the temporal evolution of fields, 
 ii) the data interchange among agents, and 
 iii) the spatial partitioning of computation. 
%
In the rest, we will present aggregate computing through the scala framework ScaFi~\cite{DBLP:journals/softx/CasadeiVAP22}.
In this variant, the three main operator constructs are \texttt{rep}, \texttt{nbr}, and \texttt{foldhood}. 
 For instance, to model the temporal evolution of a field, 
 one can employ the \lstinline[language=scafi]|rep| construct as follows:
\begin{lstlisting}[language=scafi]
// signature: def rep[A](init: A)(f: A => A): A
rep(0)(x => x + 1)
\end{lstlisting}
Where 0 is the initial value of the field, 
 and \lstinline[language=scafi]|x => x + 1| is a lambda that increments the field value by one at each cycle.
Hence, \lstinline[language=scafi]|rep| is the incremental evolution of a field with each cycle,
 representing a non-uniform field. 
 Particularly, 
 the above-described code express a field of local counters, 
 where each agent increments its own counter at each cycle.

To facilitate data exchange between agents, 
 ScaFi leverages the \lstinline[language=scafi]|nbr| construct in conjunction with a folding operator:
\begin{lstlisting}[language=scafi]
// signature: def nbr[A](value: A): A
// signature: def foldhood[A](init: A)(op: (A, A) => A)(query: A): A
foldhood(0)(_ + _)(nbr(1))
\end{lstlisting}
This snippet computes each agent's neighbor count: if agent A has 3 neighbors, 
the result will be 3, as it sums the values of its neighbors (1 in this case).
Particularly, here, \lstinline[language=scafi]|nbr(1)| captures neighboring values through a bidirectional communication primitive (namely, the agent sends the value 1 to its neighbors and receives their values in return),
\lstinline[language=scafi]|_ + _| serves as the aggregation function (summation operator, equivalent to \lstinline[language=scafi]|(a, b) => a + b|), 
and \lstinline|0| provides the initial value for the reduction operation. 
 
Lastly, to express spatio-temporal evaluation, a combination of the aforementioned constructs is utilized:
\begin{lstlisting}[language=scafi]
rep(mid) { minId => foldhood(0)(math.min)(nbr(minId)) }
\end{lstlisting}
This code snippet demonstrates a sophisticated interplay between temporal evolution and spatial aggregation to compute the global minimum identifier across the entire network.
The computation operates as follows: initially, each agent starts with its local identifier (\lstinline[language=scafi]|mid|).
At each execution round, the \lstinline[language=scafi]|rep| construct maintains the current minimum value discovered so far, while \lstinline[language=scafi]|nbr(minId)| enables each agent to share its current minimum with its neighbors.
The \lstinline[language=scafi]|foldhood| operation then aggregates these neighbor values using the \lstinline[language=scafi]|math.min| function, effectively propagating the smallest identifier throughout the network.
Through this iterative process, the minimum value gradually diffuses across the network topology: agents with smaller identifiers influence their neighbors, and this influence cascades through the network until all agents converge to the global minimum.
The temporal aspect ensures convergence stability, while the spatial aspect enables information propagation across network boundaries.
%
%
\subsubsection{Coordination Building Blocks.}
On top of this minimal set of constructs, 
 ScaFi provides a set of building blocks for developing complex coordination algorithms.
These blocks are \emph{self-stabilizing}: 
they converge to a stable state despite transient changes in network topology or agent behaviors.
A cornerstone among these constructs is the \emph{self-healing gradient} computation, 
a distributed algorithm for maintaining the minimum distance from a designated source node (\eg{} the leader) 
to every other node in the network. 
Building upon standard gradient-based approaches, 
this mechanism automatically recomputes and updates distance estimates whenever changes occur in the network (\eg{} node arrivals/removals or communication failures),
highlighting the algorithm's \emph{self-healing} property and making it highly suitable for dynamic, large-scale environments.
 Within ScaFi, this is described as follows:
\begin{lstlisting}[language=scafi]
def gradient(source: Boolean): Double =
  rep(Double.MaxValue) { dist =>
    mux(source) { 0.0 }
                {foldhood(dist) { math.min }(nbr(dist) + nbrRange())
  }
\end{lstlisting}
Where the \lstinline[language=scafi]|rep| construct maintains a distance estimate that evolves over time, 
while \lstinline[language=scafi]|mux| conditionally sets the distance to zero for source nodes.
For non-source nodes, \lstinline[language=scafi]|foldhood| aggregates neighbor distances using the minimum function,
where each neighbor's distance is incremented by the communication range (\lstinline[language=scafi]|nbrRange()|).
This creates a distributed shortest-path computation that automatically adapts to network topology changes.
Building upon this foundation,
 more sophisticated coordination algorithms can be developed, such as:
\begin{itemize}
    \item \textbf{Gradient cast}: a mechanism to disseminate information from a source node across the system using a gradient-based approach. 
    In ScaFi, this is expressed as:
    \begin{lstlisting}[language=scafi]
G[V](source: Boolean, value: V, accumulator: V => V)
    \end{lstlisting}
    Here, \lstinline[language=scafi]|source| identifies which nodes serve as the gradient's origin points, 
    \lstinline[language=scafi]|value| represents the information to be disseminated from source nodes throughout the network, 
    and \lstinline[language=scafi]|accumulator| defines how the propagated value is transformed as it spreads from node to node.
    For example, \lstinline[language=scafi]|G[Int](sense("source"), 0, _ + 1)| creates a hop-distance field that increments by one at each node as it propagates away from the source, where \lstinline[language=scafi]|sense("source")| identifies the source nodes.
    \item \textbf{Collect cast}: conceptually the inverse of gradient cast, 
    it aggregates information from the system back to a specific zone. 
    It is represented as:
\begin{lstlisting}[language=scafi]
C[V](potential: Double, accumulator: V, localValue: V, null: V)
\end{lstlisting}
Here, \lstinline[language=scafi]|potential| defines the collection gradient that establishes a spanning tree for routing values to collection points (typically following shortest paths). 
The \lstinline[language=scafi]|accumulator| parameter specifies how values are aggregated as they flow toward the collection point, 
\lstinline[language=scafi]|localValue| represents each node's contribution to the collection, and 
\lstinline[language=scafi]|null| serves as the default value when no data is available.
For example, \lstinline[language=scafi]|C(gradient(source), _ + _, 1, 0)| collects the sum of values from all nodes in the network.
    \item \textbf{Sparse choice}: 
    a distributed mechanism for node subset selection and spatial-based leader election, expressed as:
\begin{lstlisting}[language=scafi]
S(radius: Double, metric: => Double): Boolean
\end{lstlisting}
where \lstinline[language=scafi]|radius| specifies the maximum radius within which a leader can influence other nodes and \lstinline[language=scafi]|metric| defines distance used to evaluate node proximity (\eg{} Euclidean distance).
The algorithm is ``spatial-based'' as it leverages physical distances between nodes to elect leaders: 
each leader node creates a sphere of influence with radius \lstinline[language=scafi]|radius|, 
and nodes within this radius become part of that leader's region.
\end{itemize}
By integrating these constructs, 
 it becomes possible to execute complex collective behaviors, 
 such as crowd management 
 and swarm behaviors~\cite{DBLP:journals/scp/AguzziV25}.
Furthermore, it is feasible to integrate 
 the aforementioned programming model with deep learning techniques, 
 advancing towards a paradigm known as \emph{hybrid aggregate computing}~\cite{DBLP:conf/icdcs/AguzziCV22}. 
 AC can orchestrate or enhance the learning mechanisms, 
 and conversely, learning methodologies can refine AC. 
This paper adopts the initial perspective, 
 delineating a field-based coordination strategy for FL. 
 The objective is to create a distributed learning framework that is inherently self-stabilizing and self-organizing. 

\subsubsection{Self-organizing Coordination Regions.}

\begin{figure*}
  \centering
  \begin{subfigure}{0.32\textwidth}
      \centering
      \includegraphics[width=\textwidth]{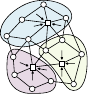}
      \caption{Information collection.}
  \end{subfigure}
  \begin{subfigure}{0.32\textwidth}
    \centering
    \includegraphics[width=\textwidth]{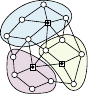}
    \caption{Leader computation.}
  \end{subfigure}
  \begin{subfigure}{0.32\textwidth}
    \centering
    \includegraphics[width=\textwidth]{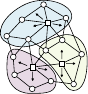}
    \caption{Information sharing.}
  \end{subfigure}
  \caption{ Graphical representation of the Self-organizing Coordination Regions pattern.  
    First, information within each area is collected in the respective leader. 
    Then, each leader processes the collected information and shares it back to the clients.
  }
  \label{fig:scr}
\end{figure*}

Recent advances in field-based coordination introduced a pattern called
Self-organizing Coordination Regions (SCR)~\cite{DBLP:conf/coordination/CasadeiPVN19}.
This strategy enables distributed collective sensing and acting without
relying on a dedicated authority, all while ensuring both self-stabilization
and self-organization of the system.
The SCR pattern operates as follows:
\begin{enumerate}
  \item A distributed multi-leader election process selects a 
   set of regional leaders across the network (\eg{} using the \texttt{S} operator);
  \item The system self-organizes into distinct regions, each governed by a single leader (\eg{} leveraging the \texttt{G} operator); 
  and
  \item Within each region, a feedback loop is established where the 
   leader collects upstream information from the agents under its influence and, 
   after processing, disseminates downstream directives (using both the \texttt{C} and \texttt{G} operators, see \Cref{fig:scr} for an overview).
\end{enumerate}

This pattern effectively combines the previously discussed building blocks 
and is straightforward to implement in the aggregate computing context.
The following ScaFi code demonstrates a compact implementation of the SCR pattern:

\begin{lstlisting}[language=scafi]
def SCR(radius: Double): Boolean = {
  val leader = S(radius)                     // Step 1: Elect leaders 
  val potential = gradient(leader)           // Step 2: Create regions
  val collectValue = 
    C(potential, accumulationLogic, localValue, nullValue) // Step 3a: Collect data
  val decision = mux(leader) {
    decide(collectValue)                     // Step 3b: Leaders process data
  } {
    nullValue
  }
  G(leader, decision, identity)              // Step 3c: Broadcast decisions
}
\end{lstlisting}

In this implementation, line 2 performs distributed leader election within the specified \lstinline[language=scafi]|radius|, 
line 3 establishes regions around each leader using gradient computation, 
lines 4-5 collect information from all agents within each region toward their respective leaders, 
lines 6-9 enable leaders to process the collected information and make decisions, 
and line 10 broadcasts these decisions back to all agents in each region.

\section{Related Works and Motivation}\label{sec:related}
In this section, we present several state-of-the-art methods that serve as baseline 
 for comparison with the proposed approach. 
Additionally, we discuss the motivation for our work 
 and describe a reference scenario that contextualizes our contribution.

\pagebreak[5]
\subsection{Baseline algorithms}\label{sec:fl}

\paragraph{\emph{FedAvg}}
One of the most common algorithms for horizontal \ac{fl} is \emph{FedAvg}, 
 where the server aggregates the models by averaging the weights of the models received from the client devices.
A \emph{model} in machine learning refers to a mathematical representation (typically a neural network in our context) that learns patterns from data to make predictions or classifications.
The \emph{weights} (also called parameters) are the numerical values within the model that are learned during training and determine how input data is transformed to produce outputs.
Formally, given a population of $K$ devices, each of them with a local dataset $D_k$ and size $n_k = |D_k|$.
Each device $k$ performs $E$ epochs of \emph{local training} on its dataset, producing an updated local model 
 $\tilde{\mathbf{w}}_k^{(t)} \in \mathbb{R}^p$ in round $t$.
The server then computes the weights of the global model $\mathbf{w}^{(t+1)}$ as a data-size weighted average of the 
 vectors from the local models shared by the devices:
\begin{equation}\label{eq:fedavg}
  \mathbf{w}^{(t+1)} = \frac{1}{\sum_{k=1}^K n_k} \sum_{k=1}^K n_k\, \tilde{\mathbf{w}}_k^{(t)}.
\end{equation}
This process is repeated for a fixed number of rounds, 
 with the server distributing the global model to the devices at the beginning of each round.

\paragraph{\emph{FedProx}}
This algorithm extends FedAvg by introducing a proximal term to the objective 
 function that penalizes large local model updates.
This modification helps address statistical heterogeneity across devices while safely 
 accommodating varying amounts of local computation due to system heterogeneity.

Formally, define the proximalized local objective for device $k$ at round $t$ as
\begin{equation}
  h_k(\mathbf{w}; \mathbf{w}^{(t)}) = F_k(\mathbf{w}) + \frac{\mu}{2} 
  \|\mathbf{w} - \mathbf{w}^{(t)}\|_2^2,
\end{equation}
where $F_k(\mathbf{w})$ is the local loss function, $\mathbf{w}^{(t)}$ is the global model at round start, 
 and $\mu \geq 0$ is the proximal term coefficient controlling how far local 
 updates can deviate from the global model.
The local update approximately solves
\begin{equation}
  \mathbf{w}_k^{(t+1)} \in \argmin_{\mathbf{w}}\; h_k(\mathbf{w}; \mathbf{w}^{(t)}).
\end{equation}

\paragraph{\emph{Scaffold}}
This algorithm introduces control variates---variance reduction techniques that track the local gradient drift---to correct the divergence of local models from the global optimum caused by heterogeneous data distributions across clients. 
Unlike FedAvg, where each client performs model updates locally and communicates its version to the server, 
 Scaffold tries to minimize this drift using control variates that track the direction of gradient descent for each client.
In the local model update phase, each client $i$ adjusts its model $y_i$ using the formula 
 $y_i \leftarrow y_i - \eta_l (g_i(y_i) + c - c_i)$, where $\eta_l$ is the local learning rate, 
 $g_i(y_i)$ represents the local gradient, and $c$ and $c_i$ are the server and client control variates respectively.
The client's control variate $c_i$ is then updated using either $c^+_i \leftarrow g_i(x)$ or 
 $c^+_i \leftarrow c_i - \frac{1}{K\eta_l}(x - y_i)$, where $x$ is the server's model 
 and $K$ represents the number of local updates.
Finally, the server performs global updates. 
The global model is updated as $x \leftarrow x + \eta_g \frac{1}{|S|} \sum_{i \in S} (y_i - x)$, 
 where $\eta_g$ is the global learning rate and $S$ is the set of selected clients. 
Simultaneously, the server control variate is adjusted using $c \leftarrow c + \frac{1}{N} \sum_{i \in S} (c^+_i - c_i)$, 
 with $N$ being the total number of clients.

\subsection{Motivation}\label{sec:motivation}

\begin{figure}[htb]
    \centering
    \includegraphics[width=0.4\columnwidth]{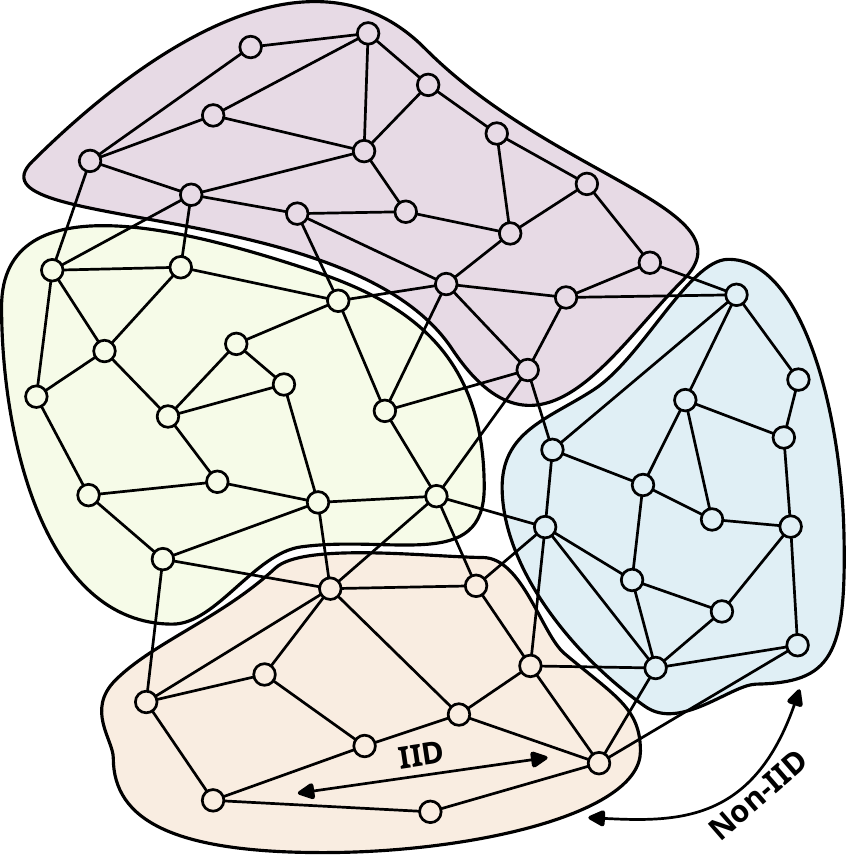}
    \caption{Spatial data distribution: homogeneous within subregions, non-IID across subregions.
    }
    \label{fig:areas}
\end{figure}

While these approaches to federate networks from multiple learning devices 
 partially address non-IID data distribution challenges, they overlook a crucial aspect: 
 the spatial distribution of devices and its relationship with data distribution patterns.
Research has shown that devices in spatial proximity often exhibit similar data distributions,
 as they typically capture related phenomena or user behaviors~\cite{DBLP:conf/acsos/DominiAFVE24}---see~\Cref{fig:areas}.
Consider, for instance, a motivating scenario involving autonomous vehicles (AVs) distributed across a metropolitan area
 collaboratively forecasting traffic conditions while preserving privacy.
AVs operating in different environments (\eg{} highways, urban centers, or suburbs) collect data with distinct characteristics,
 highlighting strong spatial autocorrelation in traffic patterns.
Collaboration among AVs in the same spatial region yields more accurate models,
 while aggregation across spatially dissimilar regions leads to degraded performance.
This spatial correlation suggests that clustering devices based on spatial proximity
 could enhance model performance and personalization.
However, current approaches in federated learning are predominantly centralized
 and rely on traditional clustering algorithms that do not consider the spatial aspects
 of data distribution.
This gap highlights the need for an approach that simultaneously addresses:
\begin{enumerate*}[label=(\roman*)]
  \item decentralization,
  \item non-\ac{iid} data handling, and
  \item spatial correlation in data distributions via distributed spatial-based leader election.
\end{enumerate*}
Our work aims to bridge this gap through field-based coordination---see \Cref{tab:methods-comparison} 
 as a comparison between current approaches and our proposal.

\begin{table}[ht]
  \centering
  \caption{Comparison of federated learning approaches.}
  \label{tab:methods-comparison}
  \begin{tabular}{lccc}
  \toprule
  	\textbf{Method} & \textbf{Decentralized} & \textbf{Non-\ac{iid}} & \textbf{Spatial Correlation} \\
  \midrule
  FedAvg   & \ding{55} & \ding{55} & \ding{55} \\
  FedProx  & \ding{55} & \ding{51} & \ding{55} \\
  Scaffold & \ding{55} & \ding{51} & \ding{55} \\
  P2P FL   & \ding{51} & \ding{55} & \ding{55} \\
  FBFL     & \ding{51} & \ding{51} & \ding{51} \\
  \bottomrule
  \end{tabular}
\end{table}

\subsection{Comparison with Other FL Variants}\label{sec:comparison}

FBFL operates within the horizontal FL paradigm, 
 and more specifically aligns with the family of personalized and clustered FL 
 approaches designed to mitigate the impact of non-IID data distributions 
 (\eg{}~\cite{DBLP:conf/nips/0001MO20,DBLP:journals/tit/GhoshCYR22,DBLP:journals/corr/abs-2505-02540}). 
In our work, the formation of self-organizing learning regions constitutes a fully distributed 
 realization of clustered FL~\cite{liu2025survey}, where spatial proximity is used as a proxy for data similarity. 
This design eliminates the need for centralized clustering procedures and allows continuous 
 reconfiguration in response to changes in network topology or data 
 distribution---a property rarely achieved in centralized clustered FL.

By contrast, vertical FL addresses scenarios where participants share the same sample space but 
 hold disjoint feature spaces~\cite{DBLP:journals/tist/YangLCT19,DBLP:journals/tkde/LiuKZPHYOZY24}, and therefore does not directly address challenges 
 such as label skew or heterogeneous sample distributions across devices---the primary focus of FBFL. 

By integrating the adaptability and personalization strengths of clustered FL with the resilience and 
 scalability of a self-organizing architecture, 
 FBFL advances the state of the art in handling spatially correlated non-IID data in fully decentralized settings.

\section{Field-Based Federated Learning}\label{sec:fbfl}

This section presents \acf{fbfl}, our core contribution that integrates aggregate computing principles 
 with federated learning to address the limitations identified in \Cref{sec:background,sec:related}.
The following subsections formalize the problem (\Cref{sec:problem-formulation}),
 detail our distributed algorithm (\Cref{sec:fbfl-algo}), 
 and provide implementation specifics (\Cref{sec:learningprocess}).
Finally, \Cref{tab:fbfl-notation} summarizes the notation used throughout this section.
\begin{table}[t]
  \centering
  \small
  \caption{Symbols used in \Cref{sec:fbfl} (Field-Based Federated Learning).}
  \label{tab:fbfl-notation}
  \begin{tabular}{lp{0.72\linewidth}}
    	\toprule
    	\textbf{Symbol} & \textbf{Meaning} \\
    \midrule
    $\mathcal{V}$ & Set of devices (nodes). \\
    $\mathcal{E}$ & Set of communication links (edges). \\
    $\mathcal{G}=(\mathcal{V},\mathcal{E})$ & Communication graph. \\
    $\mathcal{N}_d$ & Neighborhood of device $d$: $\{d'\in\mathcal{V}\mid (d,d')\in\mathcal{E}\}$. \\
    $\mathcal{D}_d$ & Local dataset on device $d$; $n_d = |\mathcal{D}_d|$ is its size. \\
    $x_i\in\mathbb{R}^m,\; y_i\in\mathcal{Y}$ & Feature vector and label of a sample; $m$ is feature dimension. \\
    $\mathbb{P}_d(x,y)$ & Local data distribution at device $d$. \\
    $\mathbf{w}_d\in\mathbb{R}^p$ & Parameters of the model at device $d$; $p$ is number of parameters. \\
    $\mathbf{w}^{(0)}$ & Initial shared model. \\
    $\tilde{\mathbf{w}}_d^{(t)}$ & Locally updated model of device $d$ at round $t$. \\
    $\mathbf{w}_l^{(t)}$ & Regional model associated with leader $l$ at round $t$. \\
    $t\in\{0,\dots,\mathcal{T}\}$ & Federated learning round index; $\mathcal{T}$ is the maximum number of rounds. \\
    $f_d(\mathbf{w})$ & Local empirical risk on device $d$. \\
    $\ell(\cdot,\cdot)$ & Loss function (\eg{} cross-entropy). \\
    $h(\mathbf{w};x)$ & Model prediction for input $x$ under parameters $\mathbf{w}$. \\
    $\alpha_d,\; \alpha_{d,l}^{(t)}$ & Data-size weights (global and regional, respectively). \\
    $\mathcal{L}^{(t)}$ & Set of elected leaders at round $t$. \\
    $\mathcal{R}_l^{(t)}$ & Region (influence set) of leader $l$ at round $t$; $\{\mathcal{R}_l^{(t)}\}_{l\in\mathcal{L}^{(t)}}$ partitions $\mathcal{V}$. \\
    $d_{\mathcal{L}}^{(t)}$ & Leader assigned to device $d$ at round $t$. \\
  $\mathrm{dist}(d,l)$ & Neighborhood metric between devices $d$ and $l$ (\eg{} Euclidean in space or hop distance on $\mathcal{G}$). \\
    $D(\mathbb{P}_d,\mathbb{P}_{d'})$ & Statistical distance between local distributions. \\
    $\epsilon$ & Threshold for spatial-data similarity (\Cref{assumption:spatial-data-correlation}). \\
    $E$ & Number of local training epochs per round. \\
  $\mathcal{DL}(\mathrm{dist}, R)$ & Distributed leader election procedure parameterized by metric and radius. \\
    $d' \Longrightarrow_t d''$ & Forward chain (multi-hop path) from node $d'$ to $d''$ at round $t$. \\
  $R$ & Influence radius (in metric units or hops): maximum distance a leader influences. \\
    $\mathcal{A}$ & Aggregation operator (\eg{} data-size weighted average). \\
    \bottomrule
  \end{tabular}
\end{table}

  
\subsection{Problem Formulation}\label{sec:problem-formulation}

Consider a distributed network comprising a set of devices, 
 denoted as $\mathcal{V}$.
In our field-based approach, devices interact through direct peer-to-peer communication rather than relying on a central server. This network can be modelled as a graph $\mathcal{G} = (\mathcal{V}, \mathcal{E})$,
 with nodes representing devices and edges symbolizing direct communication links. 
 The neighbors of a device $d$, with which $d$ can exchange messages, 
 are denoted by $\mathcal{N}_d = \{d' \in \mathcal{V} \mid (d, d') \in \mathcal{E}\}$.
\begin{assumption}[Network Connectivity]\label{assumption:network-connectivity}
The communication graph $\mathcal{G} = (\mathcal{V}, \mathcal{E})$ is connected, ensuring field-based coordination can eventually propagate information between any two devices.
\end{assumption}
\begin{assumption}[Reliable Message Delivery]\label{assumption:reliable-message-delivery}
Communication between neighboring devices $d, d' \in \mathcal{V}$ where $(d, d') \in \mathcal{E}$ is reliable, though messages may experience bounded delays.
\end{assumption}

Each device $d \in \mathcal{V}$ possesses a local dataset $\mathcal{D}_d = \{(x_i, y_i)\}_{i=1}^{n_d}$ 
where $x_i \in \mathbb{R}^m$ represents the input features, $y_i \in \mathcal{Y}$ denotes the corresponding labels, 
and $n_d = |\mathcal{D}_d|$ is the number of samples on device $d$. 
The local data distribution $\mathbb{P}_d(x,y)$ may vary across devices, 
particularly in non-\ac{iid} scenarios where $\mathbb{P}_d \neq \mathbb{P}_{d'}$ for $d \neq d'$.

The dataset on device $d$ is used to train a local model, characterized by a weights vector $\mathbf{w}_d \in \mathbb{R}^p$, 
where $p$ represents the total number of parameters in the neural network architecture. 
\begin{assumption}[Shared Model Initialization]\label{assumption:shared-model-initialization}
All devices $d \in \mathcal{V}$ start with the same initial model $\mathbf{w}_d^{(0)} = \mathbf{w}^{(0)}$, 
ensuring a common starting point for the learning process.
This may be a pre-trained model or a randomly initialized model shared via a global elected leader.
\end{assumption}
\begin{assumption}[Synchronized Federated Learning Rounds]\label{assumption:synchronized-fl-rounds}
There exists a global synchronization mechanism that ensures all devices progress through federated learning rounds in discrete time steps $t \in \{0, 1, 2, \ldots, \mathcal{T}\}$, 
where devices complete local training, model sharing, and aggregation phases within bounded time intervals before proceeding to round $t+1$.
\end{assumption}
At federated learning round $t$, the local objective function $f_d: \mathbb{R}^p \rightarrow \mathbb{R}$ measures the empirical risk on device $d$'s local dataset:
\begin{equation}
f_d(\mathbf{w}_d^{(t)}) = \frac{1}{n_d} \sum_{i=1}^{n_d} \ell(h(\mathbf{w}_d^{(t)}; x_i), y_i),
\end{equation}
where $h(\mathbf{w}_d^{(t)}; x_i)$ denotes the model's prediction for input $x_i$ given parameters $\mathbf{w}_d^{(t)}$, 
and $\ell(\cdot, \cdot)$ represents the loss function (\eg{} cross-entropy for classification tasks).

In traditional federated learning, the global objective seeks to minimize a weighted combination of all local objectives:
\begin{equation}
  f(\mathbf{w}) = \sum_{d \in \mathcal{V}} \alpha_d f_d(\mathbf{w}),
\end{equation}
where $\alpha_d = \frac{n_d}{|D|}$ represents the relative weight of device $d$'s data contribution, 
with $|D| = \sum_{d \in \mathcal{V}} n_d$ being the total number of samples across all devices.

In contrast, our field-based federated learning approach creates multiple specialized regional models through distributed spatial-based leader election. 
Let $\mathcal{L}^{(t)} \subseteq \mathcal{V}$ denote the set of elected leaders at round $t$, where each leader $l \in \mathcal{L}^{(t)}$ governs a region $\mathcal{R}_l^{(t)} \subset \mathcal{V}$ of nearby devices, and $\{\mathcal{R}_l^{(t)}\}_{l \in \mathcal{L}^{(t)}}$ forms a partition of $\mathcal{V}$.
Rather than optimizing a single global model, we minimize regional objectives at each round:
\begin{equation}
  f_l^{(t)}(\mathbf{w}_l) = \sum_{d \in \mathcal{R}_l^{(t)}} \alpha_{d,l}^{(t)} f_d(\mathbf{w}_l),
\end{equation}
where $\alpha_{d,l}^{(t)} = \frac{n_d}{\sum_{d' \in \mathcal{R}_l^{(t)}} n_{d'}}$ represents the weight of device $d$'s data within region $l$ at round $t$ 
(note that $\sum_{d \in \mathcal{R}_l^{(t)}} \alpha_{d,l}^{(t)} = 1$), 
and $\mathbf{w}_l$ is the regional model optimized for devices in $\mathcal{R}_l^{(t)}$. 
\begin{assumption}[Spatial-Data Correlation]\label{assumption:spatial-data-correlation}
Devices in spatial proximity exhibit similar data distributions, \ie{} for devices $d, d' \in \mathcal{R}_l^{(t)}$ within the same region $l$, 
the statistical distance between their local distributions satisfies $D(\mathbb{P}_d, 
\mathbb{P}_{d'}) < \epsilon$ for some threshold $\epsilon > 0$ (for some statistical distance $D$, \eg{} total variation or Wasserstein), 
while devices across different regions may have significantly different distributions.
\end{assumption}

\paragraph{\emph{Discussion of Modeling Assumptions}}
The theoretical framework of FBFL rests on several key assumptions that warrant further discussion 
 regarding their practical implications.
Regarding~\Cref{assumption:network-connectivity}, 
 strict, permanent full connectivity is not a prerequisite for the system's operation. 
As long as a communication path exists between any pair of devices, 
 field-based coordination mechanisms ensure that information eventually propagates across the network, 
 with convergence occurring in $O(D)$ time, where $D$ denotes the network diameter. 
Concerning~\Cref{assumption:reliable-message-delivery}, 
 although our experimental evaluation does not explicitly consider message losses, 
 prior work~\cite{DBLP:journals/lmcs/AguzziCV25} on aggregate computing and its foundational building blocks has extensively studied unreliable communication settings, 
 demonstrating robust collective behavior even under severe message drop rates (up to 70\%).
\Cref{assumption:shared-model-initialization} can be satisfied in practice through a lightweight initialization phase: 
 a preliminary gossip-based consensus step is sufficient to align devices 
 on a shared initial model before the learning process starts.
Finally, 
 \Cref{assumption:spatial-data-correlation} reflects empirical evidence observed in many real-world scenarios, 
 where spatial proximity induces correlations in local data distributions, 
 yielding approximately i.i.d.\ data within regions and heterogeneous distributions across regions. 
While purely spatial partitioning may sometimes be overly restrictive, 
 recent extensions of this line of work~\cite{DOMINI2026101841} propose adaptive notions of \emph{space fluidity}, 
 in which initially spatial regions are dynamically refined--expanded or contracted--based on additional metrics, 
 such as proxies for data similarity.

\subsection{Algorithm}\label{sec:fbfl-algo}
Our implementation follows a semi-centralized approach where a subset of nodes are dynamically elected as \emph{aggregators}---specialized devices responsible for model aggregation within their regions.
This hierarchical structure is self-organizing: elected leaders adapt to network topology changes (\eg{} node failures or mobility), 
 ensuring resilient coordination without manual intervention.
To understand the algorithm's operation, it is essential to formalize the concept of \emph{forward chain}:

Each node belongs to exactly one leader's region and communicates its model exclusively with that designated leader. 
We employ a distance-based leader selection rule that establishes a Voronoi-like partitioning of the network.
Formally, given a node $d$ and the leader set $\mathcal{L}^{(t)} \subseteq \mathcal{V}$ computed by the distributed leader election algorithm $\mathcal{DL}(\mathrm{dist}, R)$ at round $t$, 
with an application-chosen influence radius $R>0$ and a time-independent neighborhood metric $\mathrm{dist}(\cdot,\cdot)$ (\eg{} Euclidean in physical space or hop-count over $\mathcal{G}$),
node $d$ falls under the influence of leader $l \in \mathcal{L}^{(t)}$ if and only if:
\begin{equation}
\mathrm{dist}(d,l) \le R \quad \text{and} \quad \forall \, l' \in \mathcal{L}^{(t)}\setminus\{l\}, \; \mathrm{dist}(d,l) < \mathrm{dist}(d,l'),
\end{equation}
Note that, 
for every node $d$ there exists a leader $l$ with $\mathrm{dist}(d,l)\le R$, 
hence each node is assigned. 
In case of no leader being within radius $R$, 
the node becomes leader $d_{\mathcal{L}}^{(t)} = d$.
In cases of equidistance, 
a predefined tie-breaking rule is applied.
The set of nodes under leader $l$'s influence is denoted $\mathcal{R}_l^{(t)}$, 
while $d_{\mathcal{L}}^{(t)}$ represents the leader to which node $d$ is assigned at round $t$.
A forward chain from node $d'$ to node $d''$ 
at round $t$ (denoted $d' \Longrightarrow_t d''$) 
is defined as the sequence, constrained to remain within the leader's influence radius when $d''$ is a leader:
\begin{equation}
d_1, d_2, \ldots, d_k \quad \text{such that} \quad d_1 = d', \quad d_k = d'', \quad \text{and} \quad (d_i, d_{i+1}) \in \mathcal{E} \; \text{ for } i=1,\ldots,k-1,
\end{equation}
with $\mathcal{E}$ possibly depending on $t$ in time-varying topologies. 
This formalization enables communication between nodes lacking direct connectivity 
through a chain of intermediate connections; 
in practice, routing follows the gradient-induced minimal-potential paths computed by the coordination layer.

Given these definitions, the round-$t$ evolution is as follows (see \Cref{algo:sc-fl} for a comprehensive overview):
\begin{enumerate}
  \item \emph{Leader Election:} A distributed leader election algorithm $\mathcal{DL}(\mathrm{dist}, R)$ 
  is executed to select a set of leaders $\mathcal{L}^{(t)} \subseteq \mathcal{V}$ 
  and induce regions $\{\mathcal{R}_l^{(t)}\}_{l \in \mathcal{L}^{(t)}}$. 
\begin{assumption}[Gradual Topology Changes]\label{assumption:gradual-topology-changes}
Network topology and leader set changes occur at a rate slower than field computation convergence, \ie{} 
 $T_{\text{topology}} \gg T_{\text{field}}$, where $T_{\text{field}}$ and $T_{\text{topology}}$ denote 
 field stabilization time and average time between topology changes, respectively.
\end{assumption}

  \item \emph{Local Training:} Each device $d \in \mathcal{V}$ starts from $\mathbf{w}_d^{(t)}$ and performs local training on $\mathcal{D}_d$ for $E$ epochs, 
  producing an updated local model $\tilde{\mathbf{w}}_d^{(t)}$ by (approximately) minimizing $f_d(\cdot)$.
  \item \emph{Leader Selection:} Each device $d$ determines its leader $d_\mathcal{L}^{(t)} \in \mathcal{L}^{(t)}$ 
  based on its distance to the leaders, constrained by radius $R$; 
 following the partitioning rule described above.
  \item \emph{Model Sharing:} Each device $d$ shares its local model $\tilde{\mathbf{w}}_d^{(t)}$ with its leader $d_\mathcal{L}^{(t)}$ along the forward chain $d \Longrightarrow_t d_\mathcal{L}^{(t)}$, which by construction lies within radius $R$ at steady state.
  \item \emph{Model Aggregation:} Each leader $l \in \mathcal{L}^{(t)}$ 
  aggregates the models from all devices in its influence region $\mathcal{R}_l^{(t)}$ 
  using an aggregation function $\mathcal{A}$, 
 computing the new regional model weights for the next round:
\begin{equation}
  \mathbf{w}_l^{(t+1)} = \mathcal{A}\!\left(\mathbf{w}_l^{(t)}, \left\{ \big(\tilde{\mathbf{w}}_{d}^{(t)}, n_d\big) \mid d \in \mathcal{R}_l^{(t)} \right\}\right)
  \quad \text{\eg{}} \quad
  \mathbf{w}_l^{(t+1)} = \sum_{d \in \mathcal{R}_l^{(t)}} \alpha_{d,l}^{(t)} \tilde{\mathbf{w}}_{d}^{(t)}.
\end{equation}
\begin{assumption}[Aggregation Capacity Constraints]\label{assumption:aggregation-capacity-constraints}
Each elected leader $l \in \mathcal{L}^{(t)}$ possesses sufficient computational and communication resources 
 to aggregate models from all devices in its influence region $\mathcal{R}_l^{(t)}$ within the available time window 
 of round $t$.
\end{assumption}
  \item \emph{Model Dissemination:} Each leader $l$ disseminates $\mathbf{w}_l^{(t+1)}$ to all devices in its influence region $\mathcal{R}_l^{(t)}$ (bounded by $R$) following $l \Longrightarrow_t d'$ for all $d' \in \mathcal{R}_l^{(t)}$, and devices update
\begin{equation}
  \forall d \in \mathcal{R}_l^{(t)}:\quad \mathbf{w}_d^{(t+1)} \coloneqq \mathbf{w}_l^{(t+1)}.
\end{equation}
\end{enumerate}
Steps 1--6 are repeated for $t=0,1,\ldots,\mathcal{T}-1$, 
 allowing regional models to evolve iteratively based on local training and field-coordinated aggregation.

\begin{algorithm}[t]
  \caption{Field-Based Federated Learning Algorithm}\label{algo:sc-fl}
  \begin{algorithmic}[1]
  \Require number of devices $K$, Network graph $\mathcal{G}$, Epochs per round $E$, Maximum rounds $\mathcal{T}$, Initial model $\mathbf{w}^{(0)}$
  \Procedure{FBFL}{}
    \For{$\text{round} = 0$ \textbf{to} $\mathcal{T}-1$}
  \State $\text{Leaders} \gets \mathcal{DL}(\mathrm{dist}, R)$ \Comment{election at round $t$}
      \ForAll{$d \in \mathcal{V}$}
        \State \textbf{local training:} from $\mathbf{w}_d^{(t)}$ produce $\tilde{\mathbf{w}}_d^{(t)}$ on $d$'s dataset for $E$ epochs
        \State \textbf{leader selection:} Compute $d_\mathcal{L}^{(t)}$ using $\text{Leaders}$
        \State \textbf{model sharing:} Send $\tilde{\mathbf{w}}_d^{(t)}$ to $d_\mathcal{L}^{(t)}$ following $d \Longrightarrow_t d_\mathcal{L}^{(t)}$
      \EndFor
      \ForAll{$d \in \mathcal{V}$}
        \If{$d \in \text{Leaders}$}
          \State \textbf{model aggregation:} $\mathbf{w}_d^{(t+1)} \gets \mathcal{A}\big(\mathbf{w}_d^{(t)}, \{\tilde{\mathbf{w}}_{d'}^{(t)} \mid d' \in \mathcal{R}_d^{(t)}\}\big)$
          \ForAll{$d' \in \mathcal{R}_d^{(t)}$}
            \State \textbf{dissemination:} deliver $\mathbf{w}_d^{(t+1)}$ to $d'$ following $d \Longrightarrow_t d'$
          \EndFor
        \EndIf
      \EndFor
      \State Each $d' \in \mathcal{V}$ sets $\mathbf{w}_{d'}^{(t+1)}$ to the received regional model
    \EndFor
  \EndProcedure
  \end{algorithmic}
\end{algorithm}

\subsection{Implementation Details}\label{sec:learningprocess}
Our hierarchical \ac{fl} framework represents a direct application of the Self-organizing Coordination Regions (SCR) pattern~\cite{DBLP:conf/coordination/CasadeiPVN19}.
The following ScaFi implementation demonstrates this field-based coordination:

\noindent\begin{minipage}{\textwidth}
\begin{lstlisting}[language=scafi, caption={FBFL implementation using aggregate computing primitives.}]
val leaders = S(radius, nbrRange)           // Spatial-based leader election
def syncCondition: Boolean = ???              // Define synchronization condition through shared clock
rep(initModel())(localModel => {
  if(syncCondition) { localModel.trainLocally() }               // Local SGD epoch  training step
  val potential = gradient(leaders)         // Compute routing potential
  val collectedModels = C(potential, _ ++ _, Set(localModel), Set.empty)
  val aggregatedModel = if (leaders) fedAvg(collectedModels) else emptyModel
  val globalModel = G(leaders, aggregatedModel, identity)
  mux(syncCondition) { globalModel } { localModel }
})
\end{lstlisting}
\end{minipage}
\paragraph{\emph{Building Blocks Semantics.}}
\noindent This implementation leverages the aggregate computing primitives introduced in \Cref{sec:background-fbc} to provide distributed model coordination.
The formal algorithm in \Cref{algo:sc-fl} maps directly to these field-based operations:
\begin{itemize}
  \item \emph{Leader Election} (Step 1): Implemented using \lstinline[language=scafi]|S(radius, nbrRange)| for spatial-based leader selection, computing $\mathcal{L}^{(t)} = \mathcal{DL}(\text{nbrRange}, \text{radius})$.
  \sloppypar
  \item \emph{Model Sharing} (Step 4): The forward chain $d \Longrightarrow_t d_\mathcal{L}^{(t)}$ is based on collect-cast \lstinline[language=scafi]|C(potential, _ ++ _, Set(localModel), Set.empty)|, 
  where the potential field routes model updates $\tilde{\mathbf{w}}_d^{(t)}$ toward appropriate aggregators.
  \item \emph{Model Dissemination} (Step 6): Broadcast from leaders to regions $l \Longrightarrow_t d'$ via gradient-cast \lstinline[language=scafi]|G(leaders, aggregatedModel, identity)|, 
  delivering $\mathbf{w}_l^{(t+1)}$ to all $d' \in \mathcal{R}_l^{(t)}$.
\end{itemize}
\paragraph{\emph{Execution cadence}}
Devices execute the aggregate program asynchronously at frequency $f_{\text{field}}$ (Sec.~\ref{sec:background-fbc}).
\ac{fl} rounds advance when a shared-clock condition~\cite{pianini2016improving} holds \\
(\lstinline[language=scafi]|syncCondition|), typically at a lower frequency $f_{\text{FL}}$.
%
When \lstinline[language=scafi]|syncCondition| becomes true 
(entering round $t$), 
devices rely on the stabilized coordination fields (leaders and routing) to advance one federated-learning round: 
(i) they perform local training, and 
(ii) leaders aggregate a new regional model.
 Model propagation (upstream to leaders and downstream to devices) is continuous and may span multiple field-update steps; 
 it is decoupled from, and may not complete within, the same round.

\paragraph{\emph{Convergence and cost}}
Leader election and both upstream/downstream model exchange stabilize in $\mathcal{O}(D)$ field-update steps,
where $D$ is the communication-graph diameter, since gradients converge along shortest paths~\cite{mo2022automatica}.

\paragraph{\emph{Decoupling of rates}}
Running the field layer at $f_{\text{field}}\gg f_{\text{FL}}$ keeps leaders and potential fields correct,
so when \lstinline[language=scafi]|syncCondition| triggers, devices use a stabilized coordination state.
This satisfies the round-level synchronization requirements of
Assumption~\ref{assumption:synchronized-fl-rounds}.

\paragraph{\emph{Inconsistent states}}
It is also feasible to set $f_{\text{field}} = f_{\text{FL}}$, \ie{} run coordination and FL rounds at the same cadence.
This reduces the number of models exchanged during each round, 
but immediately after aggregation some nodes may temporarily rely on stale gradients/routing,
causing short-lived regional inconsistencies (e.g., adjacent nodes bound to different leaders or holding pre- vs.\ post-aggregation models)
until the fields restabilize.
We used this configuration in our experiments; 
despite these transients, the system quickly re-stabilized and showed robust performance.

\section{Experimental Evaluation}\label{sec:evaluation}

\subsection{Experimental setup}
To evaluate the proposed approach, 
 we conducted experiments on three well-known used computer vision datasets: 
 MNIST~\cite{lecun2010mnist}, FashionMNIST~\cite{DBLP:journals/corr/abs-1708-07747}, and 
 Extended MNIST~\cite{DBLP:journals/corr/CohenATS17}---their characteristics are summarized in~\Cref{tab:datasets}.
In particular, data import and partitioning were efficiently managed 
 by leveraging the ProFed benchmark~\cite{DBLP:journals/corr/abs-2503-20618}.
We employed three state-of-the-art federated learning algorithms for comparison: 
 FedAvg~\cite{DBLP:conf/aistats/McMahanMRHA17}, FedProx~\cite{DBLP:conf/mlsys/LiSZSTS20}, 
 and Scaffold~\cite{DBLP:conf/icml/KarimireddyKMRS20}.

\begin{figure*}
  \centering
  \begin{subfigure}{0.32\textwidth}
    \centering
    \includegraphics[width=\textwidth]{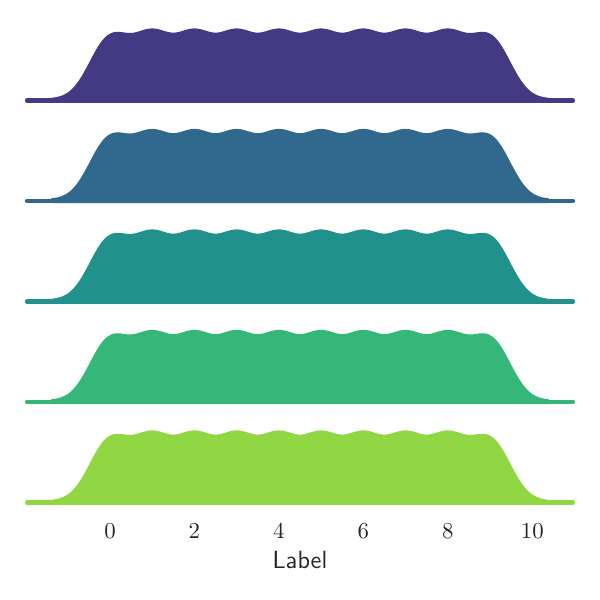}
    \caption{IID data.}
\end{subfigure}
  \begin{subfigure}{0.32\textwidth}
      \centering
      \includegraphics[width=\textwidth]{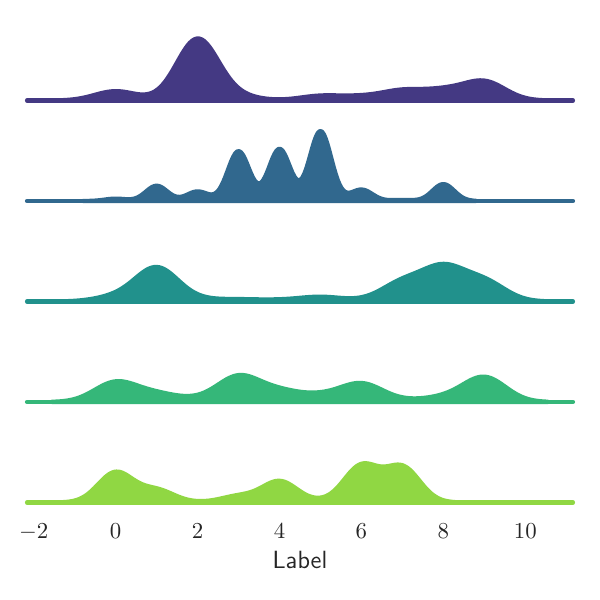}
      \caption{Dirichlet distribution.}
  \end{subfigure}
  \begin{subfigure}{0.32\textwidth}
    \centering
    \includegraphics[width=\textwidth]{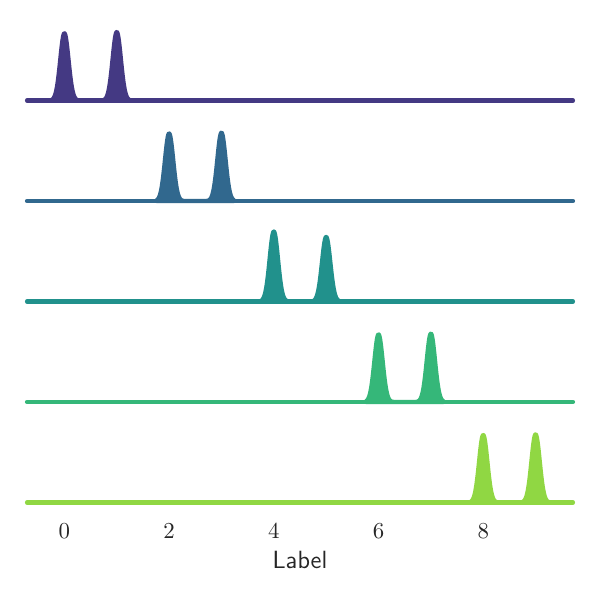}
    \caption{Hard partitioning.}
  \end{subfigure}
  \caption{ A graphical representation of three different data distribution in 5 subregions. 
  Each color represents a different subregion.
  The second and the third images are two examples of non-IID data.}
  \label{fig:distributions}
\end{figure*}

First, we evaluated FBFL against FedAvg under a homogeneous data distribution 
 to assess its stability in the absence of data skewness. 
We then created multiple heterogeneous data distributions through synthetic partitioning and 
 compared FBFL with all three baseline algorithms. 
Specifically, we employed two partitioning strategies: 
\begin{enumerate*}[label=(\roman*)]
  \item a Dirichlet-based split, where each party received samples from most classes, 
   but with a highly imbalanced distribution, leading to certain labels 
   being significantly underrepresented or overrepresented 
   (as previously done in the literature, e.g.,~\cite{DBLP:conf/nips/LinKSJ20,DBLP:conf/icml/YurochkinAGGHK19}); and
  \item a hard data partitioning strategy~\cite{DBLP:journals/apin/XuYDH25,DBLP:journals/corr/abs-2411-12377}, 
   where each region contained only a subset of labels, resulting in a more challenging learning scenario. 
\end{enumerate*}
A graphical representation of these distributions is given in~\Cref{fig:distributions}.
In particular, the dataset partitioning process used to assign data to each device followed these steps:
\begin{enumerate*}[label=(\roman*)]
  \item The dataset was first partitioned according to a predefined distribution (\eg{} Dirichlet or Hard), 
   generating a separate dataset for each subregion;
  \item Each device was then assigned to one (and only one) subregion; and
  \item For each device, a subset of data was randomly sampled from the dataset of its corresponding subregion. 
   Sampled data points were removed from the subregion's dataset to ensure that 
   no two devices shared the same data samples.
\end{enumerate*}
As a result, the data assigned to devices within the same subregion are IID, 
 while those assigned to devices across different subregion are non-IID (\Cref{fig:areas}).
Experiments were conducted using different numbers of regions, specifically $A \in \{3, 6, 9\}$,
 and $50$ devices.

Finally, we assessed the resilience of our self-organizing hierarchical architecture by simulating 
 a failure scenario. 
In this experiment, we considered a setup with four subareas and initially allowed 
 our proposed learning process to run as intended.
After a predefined amount of time---specifically, after 10 global communication rounds---we simulated 
 the failure of two out of the four leader nodes. 
The failure model was designed by randomly selecting two leader nodes and severing their communication 
 links with all other nodes in the system, effectively isolating them.
This experiment was intended to evaluate whether the system could self-stabilize by re-electing 
 new leader nodes to function as aggregator servers, 
 thereby effectively restoring a stable federation configuration aligned with the subarea structure.

All experiments were implemented using PyTorch~\cite{paszke2017automatic} for training.
Additionally, the proposed approach leverages ScaFi for aggregate computing and 
 ScalaPy~\cite{DBLP:conf/scala/LaddadS20} to enable interoperability between ScaFi and Python.
Moreover, we used a well-known simulator for pervasive systems, namely: Alchemist~\cite{DBLP:journals/jos/PianiniMV13}.
To ensure robust results, each experiment was repeated five times with different seeds.

To promote the reproducibility of the results, all the source code, 
the data, and instructions for running have been made available on 
 GitHub; both for the baselines\footnote{\url{https://github.com/davidedomini/experiments-2025-lmcs-field-based-FL-baselines}}
 and for the proposed approach\footnote{\url{https://github.com/davidedomini/experiments-2025-lmcs-field-based-FL}}.

\begin{table}
  \centering
  \caption{Overview of the datasets used in the experiments.}
  \label{tab:datasets}
  \begin{tabular}{lcccc}
    \toprule
    \textbf{Dataset} & \textbf{Training Instances} & \textbf{Test Instances} & \textbf{Features} & \textbf{Classes} \\
    \midrule
    MNIST          & 60\,000  & 10\,000 & 784 & 10 \\
    Fashion MNIST  & 60\,000  & 10\,000 & 784 & 10 \\
    Extended MNIST & 124\,800 & 20\,800 & 784 & 27 \\
    \bottomrule
  \end{tabular}
\end{table}

\subsection{Discussion}
\begin{figure*}
  \centering
  \begin{subfigure}{0.4\textwidth}
      \centering
      \includegraphics[width=\textwidth]{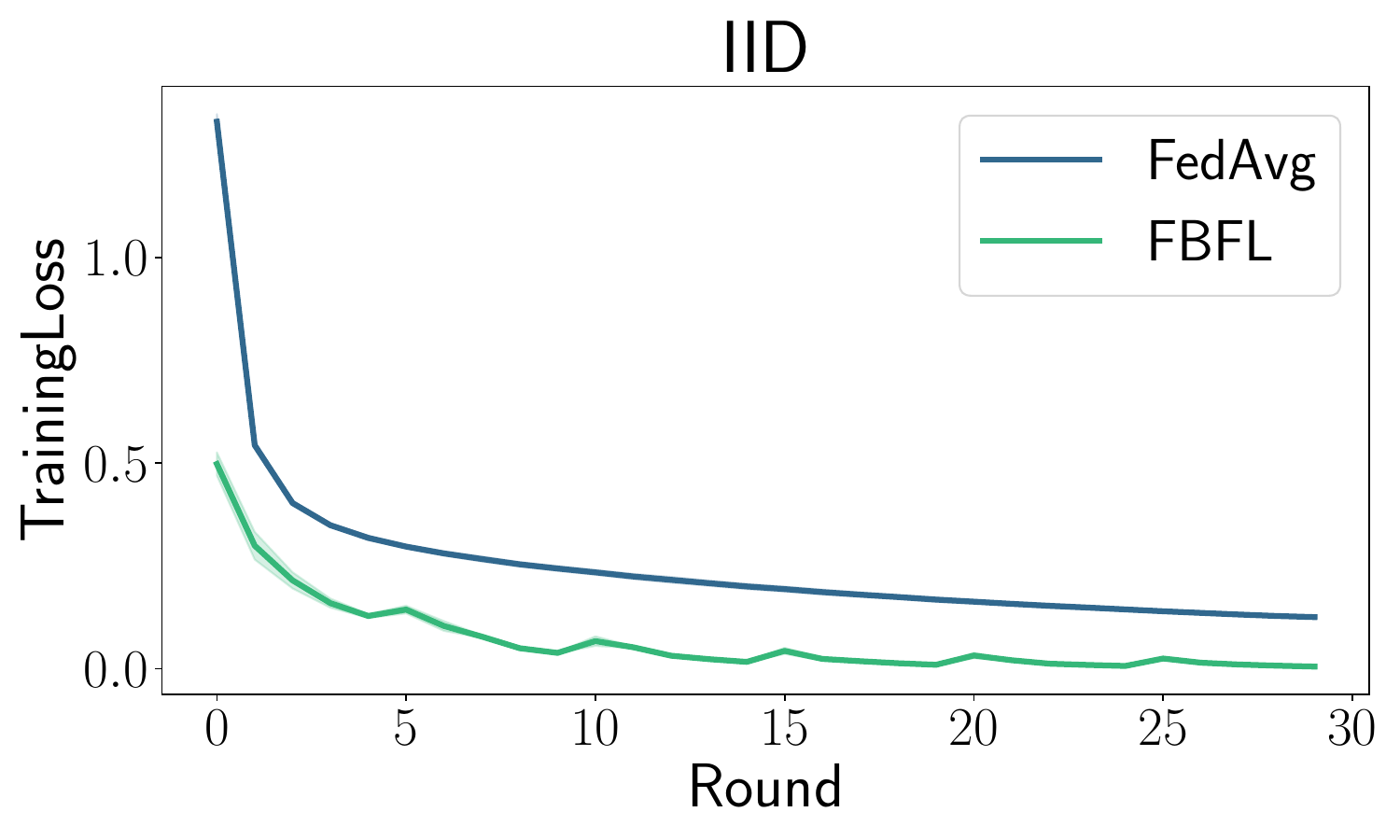}
  \end{subfigure}
  \begin{subfigure}{0.4\textwidth}
    \centering
    \includegraphics[width=\textwidth]{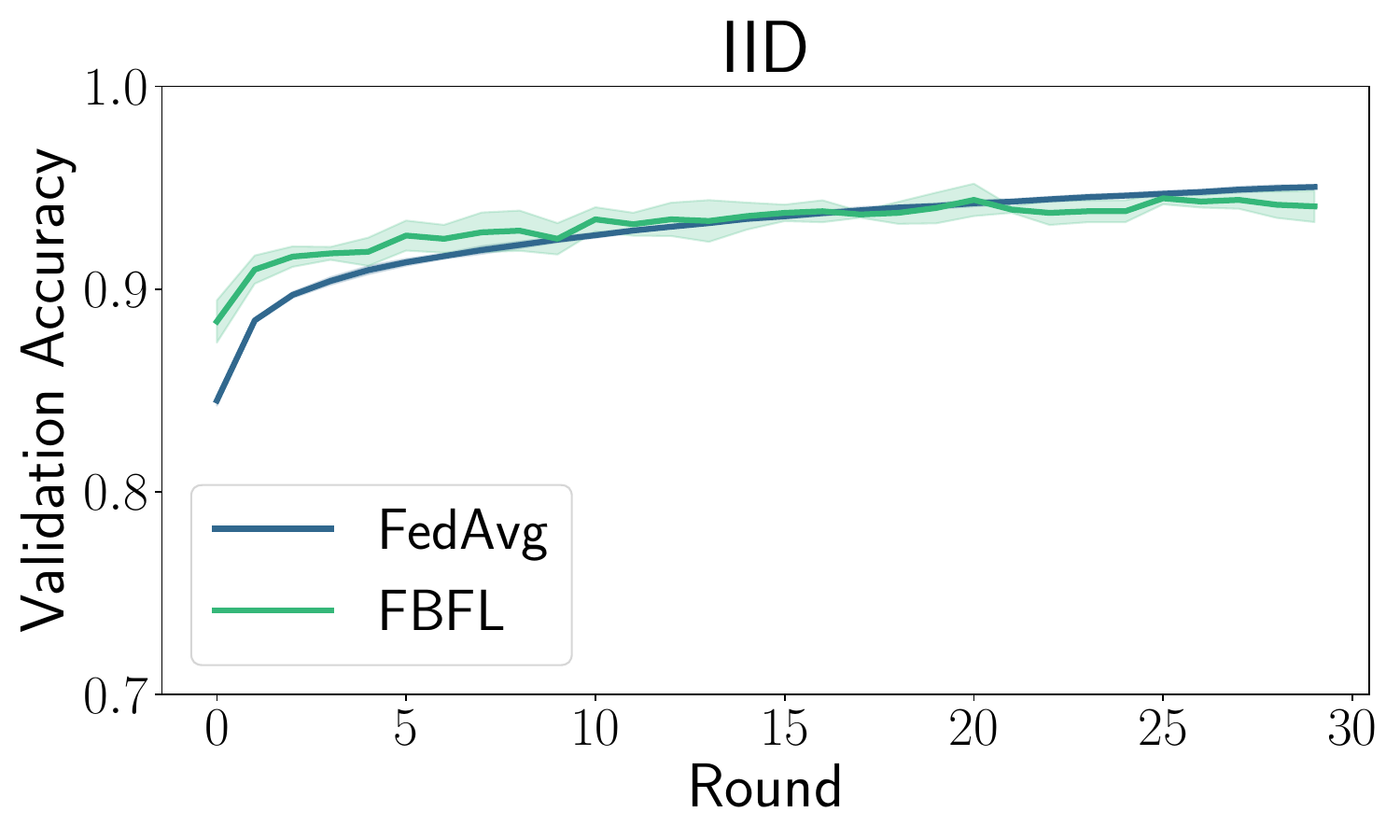}
  \end{subfigure}
  \begin{subfigure}{0.4\textwidth}
    \centering
    \includegraphics[width=\textwidth]{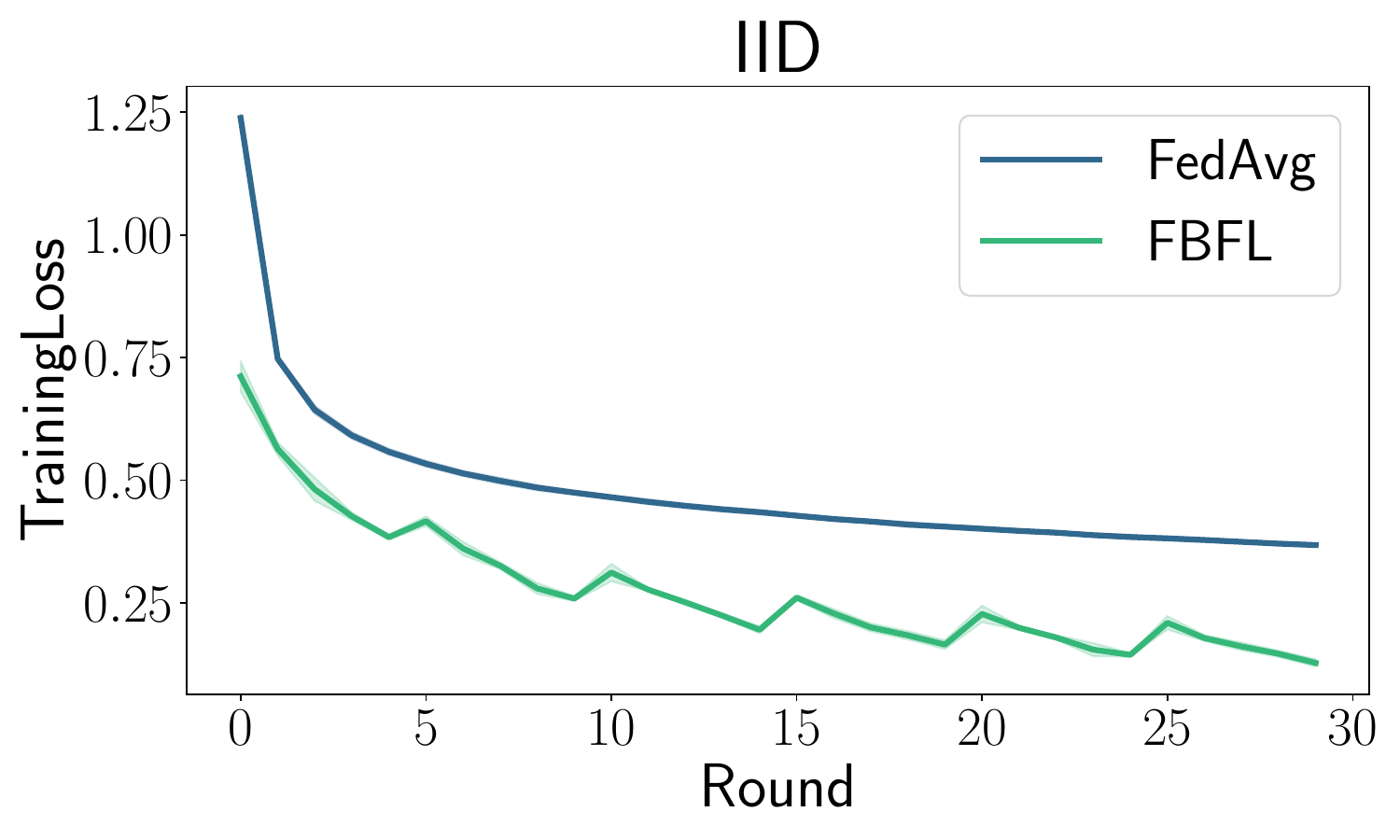}
  \end{subfigure}
  \begin{subfigure}{0.4\textwidth}
    \centering
    \includegraphics[width=\textwidth]{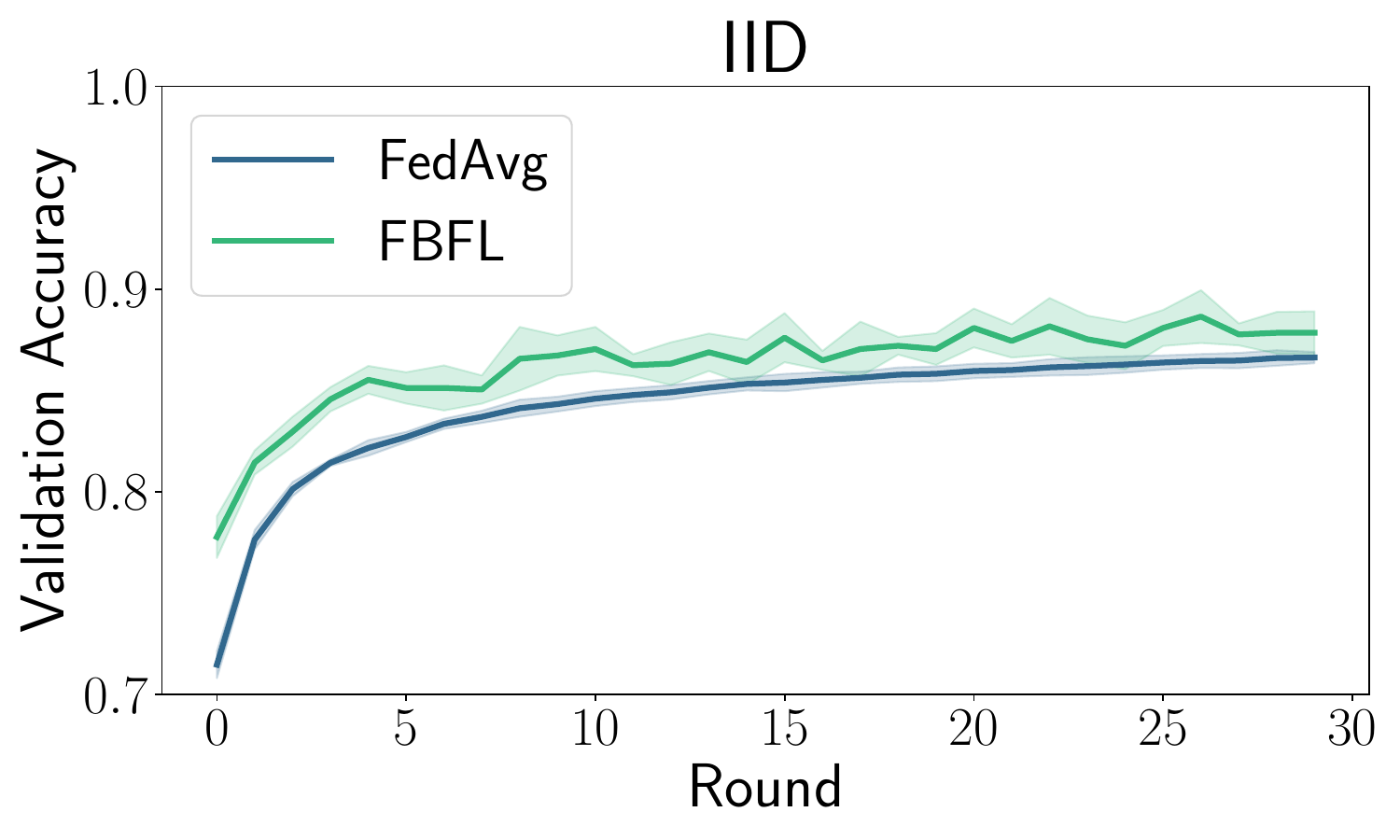}
  \end{subfigure}
  \caption{ Comparison of the proposed method (FBFL) with FedAvg under IID data. 
  The first row represents results on the MNIST dataset, while the second row on the Fashion MNIST dataset. }
  \label{fig:exp-iid}
\end{figure*}

In the following, we present the results of our experiments, 
 comparing the proposed approach with the baseline algorithms under different conditions 
 and replying to the research questions posed in~\Cref{sec:rqs}.
\begin{tcolorbox}[colback=gray!10!white, colframe=gray!80!black, title=\textbf{\ref{itm:rq1}}]
\emph{How does the Field-Based Federated Learning approach perform compared to centralized FL under IID data?}
\end{tcolorbox}
\noindent To answer~\ref{itm:rq1}, we evaluated FBFL against FedAvg under homogeneous data distribution.
The goal of this evaluation was to show how the proposed approach, based on field coordination, 
 achieves comparable performance to that of centralized FL while introducing all 
 the advantages discussed in~\Cref{sec:fbfl}, such as: the absence of a single point 
 of failure and the adaptability.
\Cref{fig:exp-iid} shows results on the MNIST (first row) and Fashion MNIST (second row) datasets. 
It is worth noting that both the training loss and the validation accuracy exhibit similar trends. 
As expected, our decentralized approach shows slightly more pronounced oscillations in both metrics
compared to FedAvg, due to the lack of a central coordinating authority.
However, despite these inherent fluctuations in the decentralized setting,
FBFL still achieves comparable final performance after the same number of global communication rounds,
matching the effectiveness of traditional centralized learning methods as observed in the literature.

\begin{figure*}
  \centering
  \begin{subfigure}{0.4\textwidth}
      \centering
      \includegraphics[width=\textwidth]{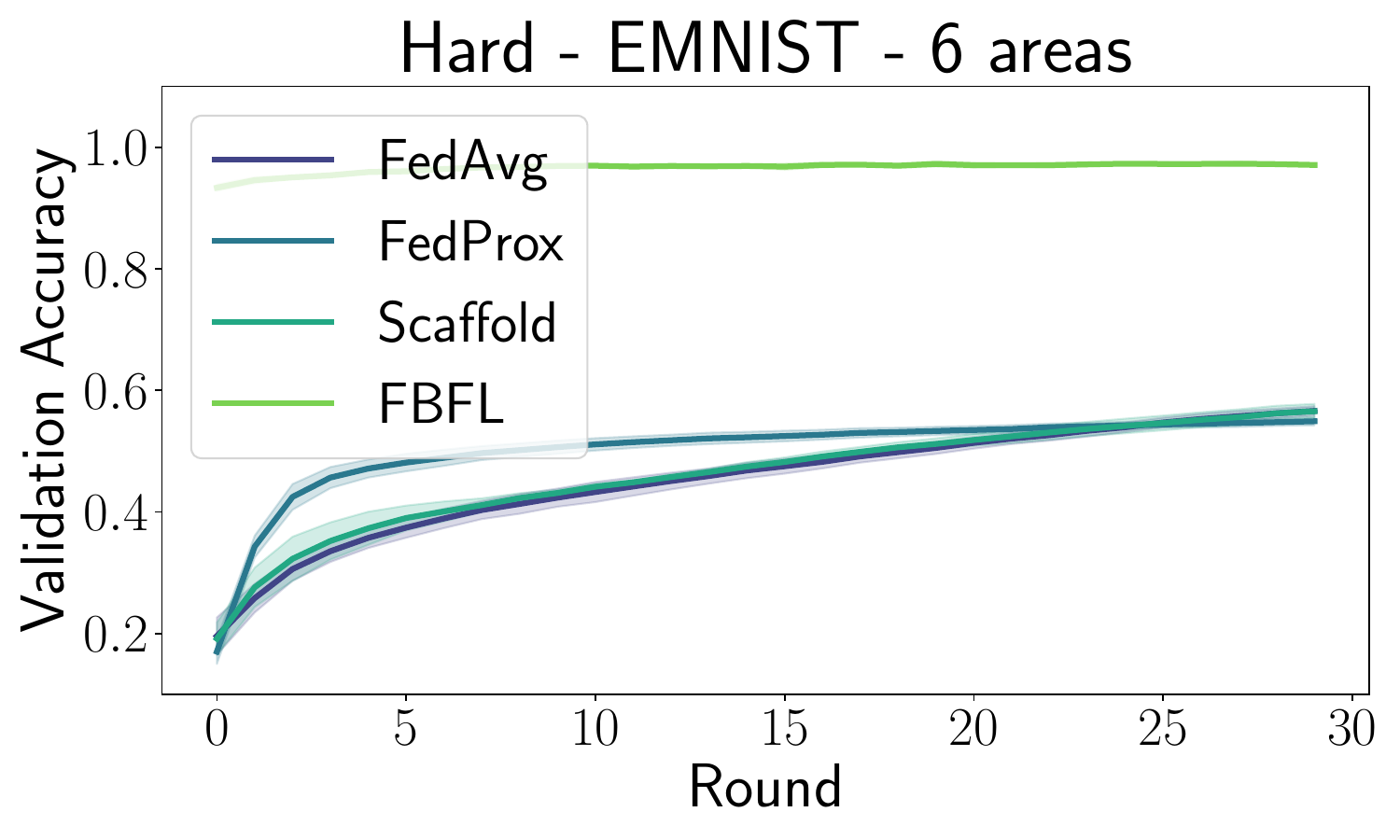}
  \end{subfigure}
  \begin{subfigure}{0.4\textwidth}
    \centering
    \includegraphics[width=\textwidth]{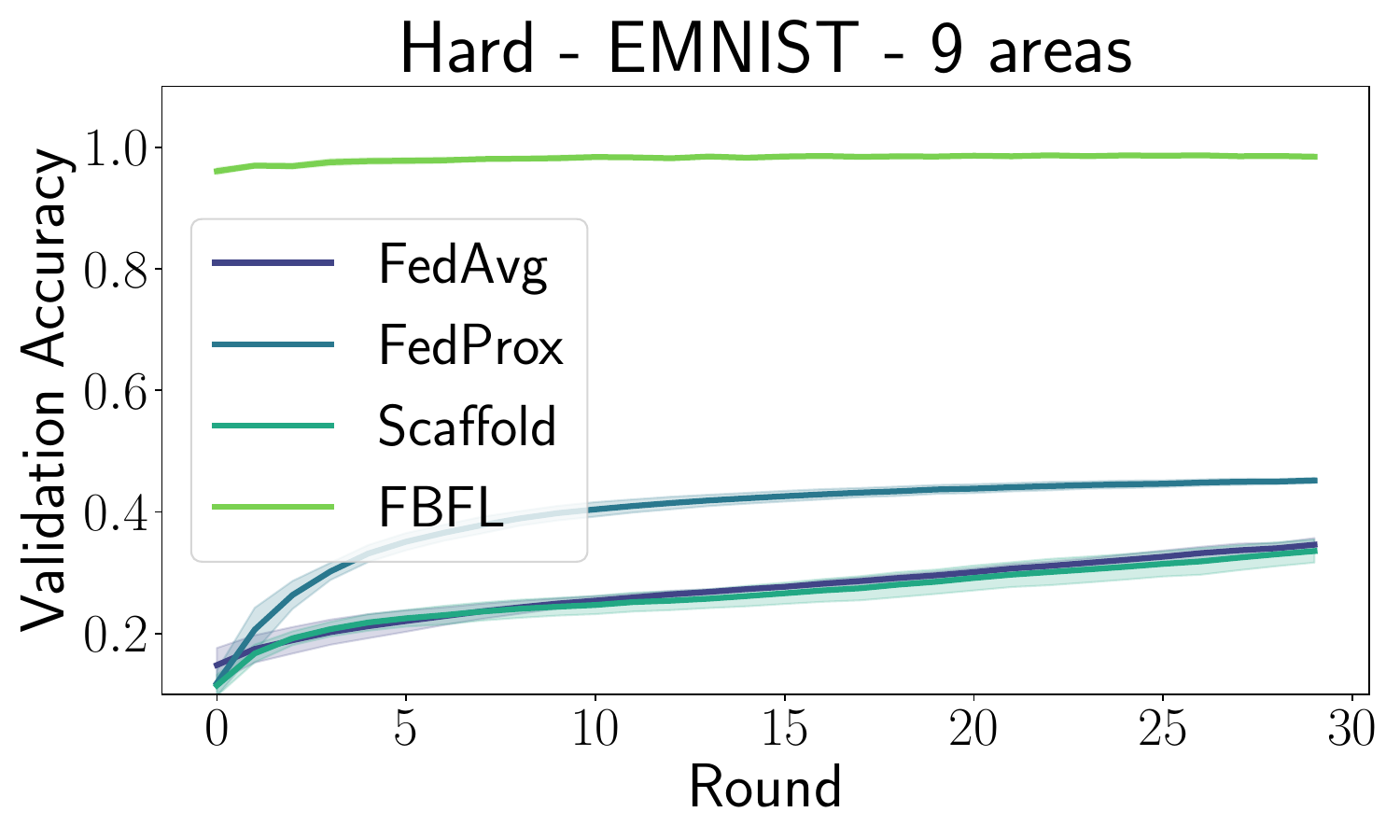}
  \end{subfigure}
  \begin{subfigure}{0.4\textwidth}
    \centering
    \includegraphics[width=\textwidth]{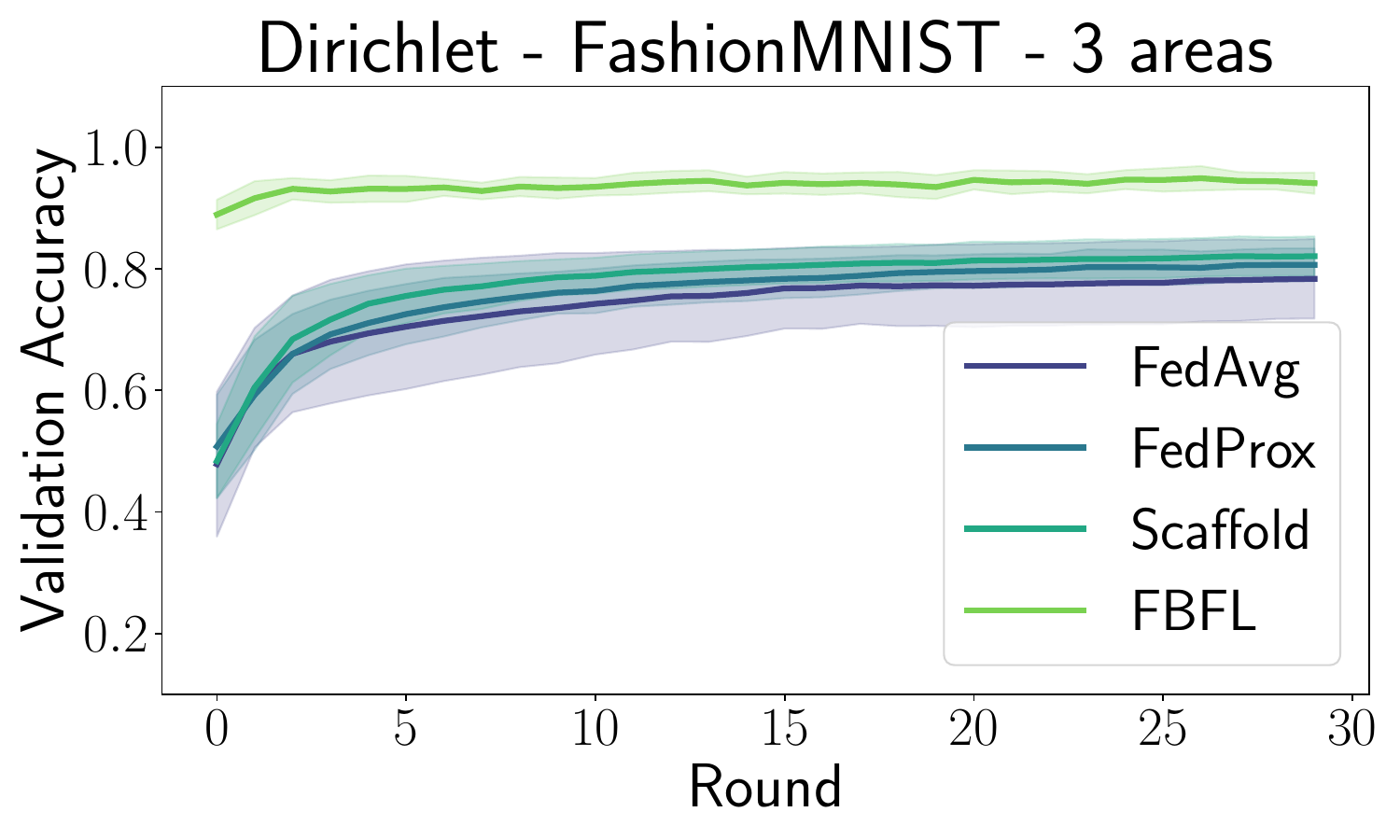}
  \end{subfigure}
  \begin{subfigure}{0.4\textwidth}
    \centering
    \includegraphics[width=\textwidth]{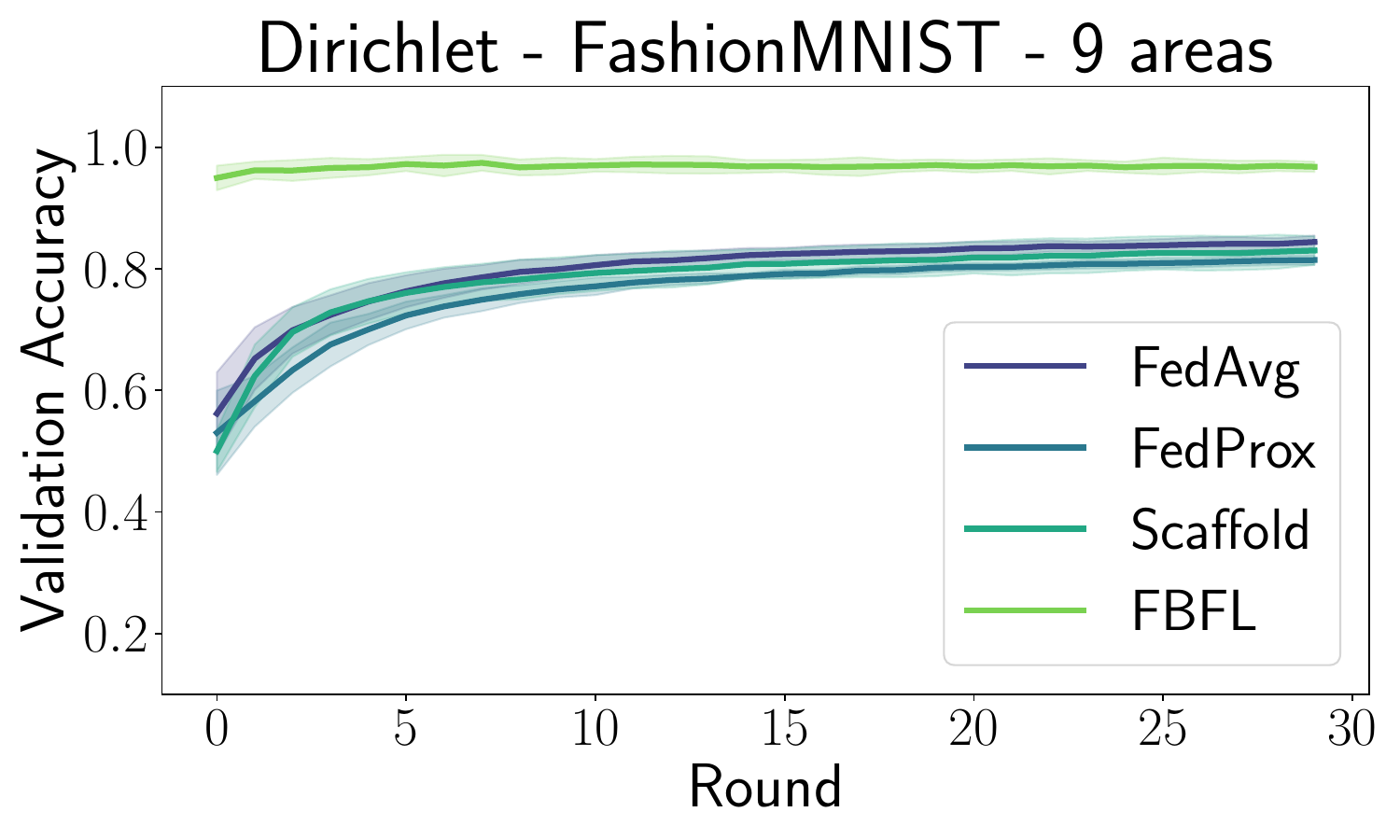}
  \end{subfigure}
  \begin{subfigure}{0.4\textwidth}
    \centering
    \includegraphics[width=\textwidth]{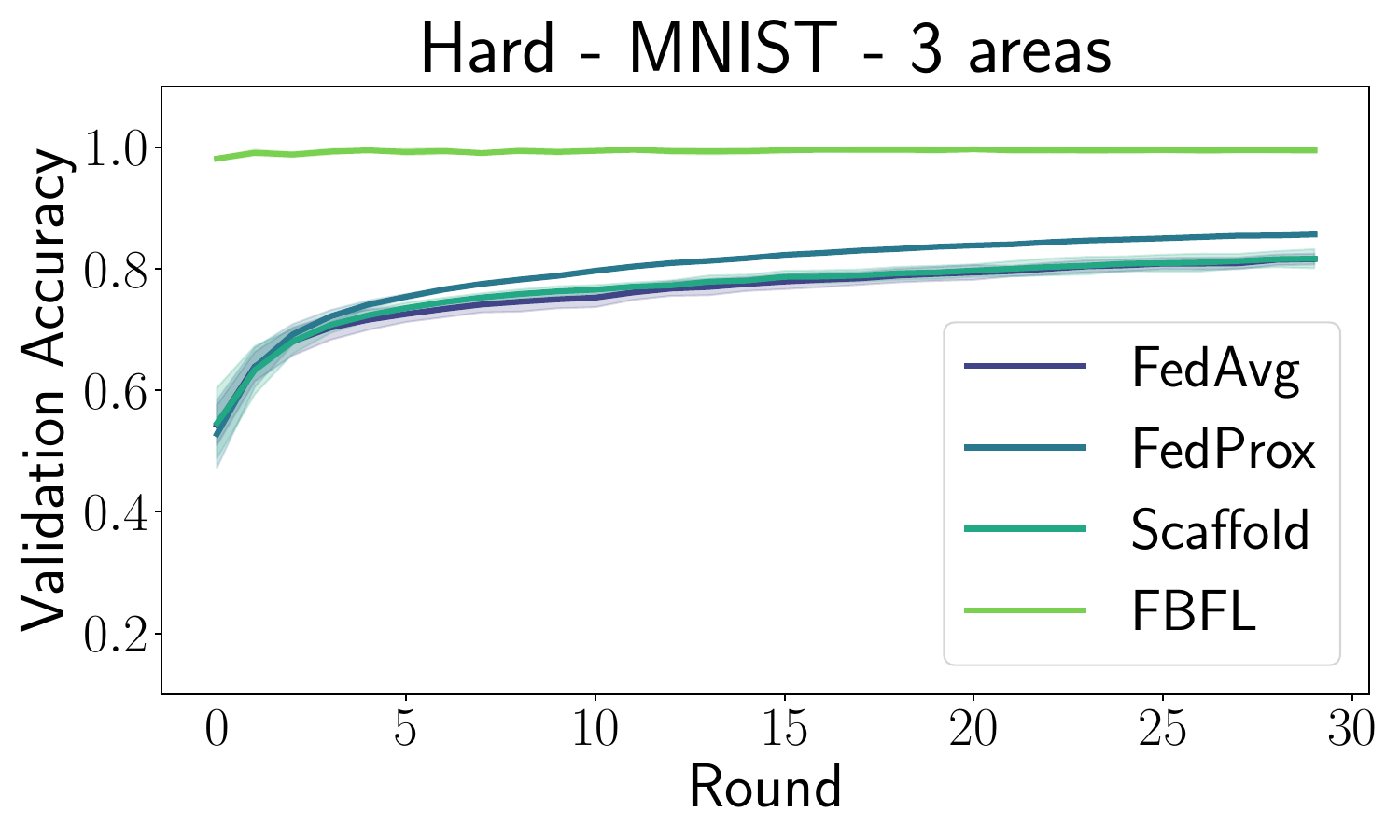}
  \end{subfigure}
  \begin{subfigure}{0.4\textwidth}
    \centering
    \includegraphics[width=\textwidth]{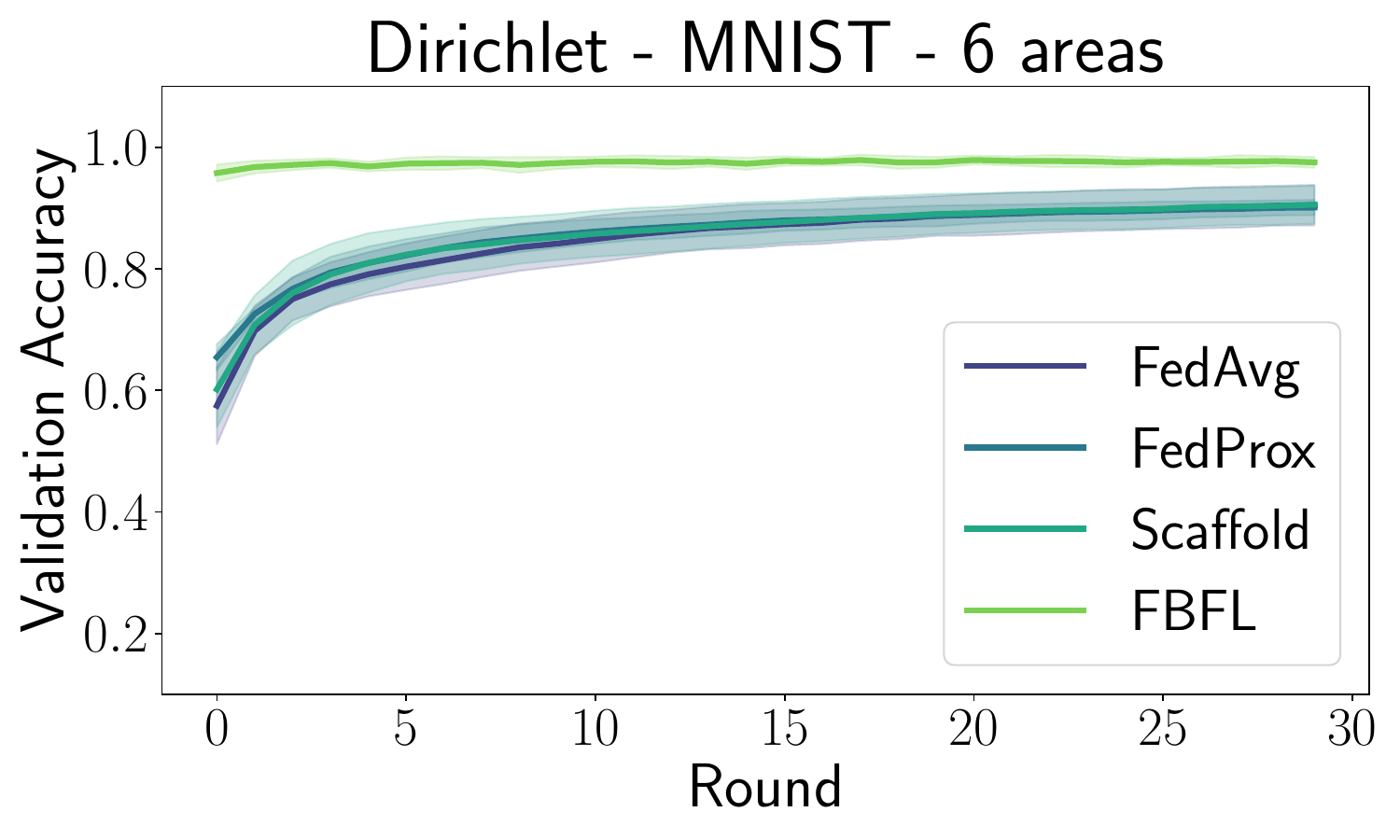}
  \end{subfigure}
  \caption{Comparison of the proposed method (FBFL) with all the baselines under non-IID data.
  The first row represents results on the Extended MNIST dataset, while the
  second row on the Fashion MNIST dataset and the third on the MNIST dataset.
  }
  \label{fig:exp-non-iid}
\end{figure*}

\begin{tcolorbox}[colback=gray!10!white, colframe=gray!80!black, title=\textbf{\ref{itm:rq2}}]
\emph{Can we increase accuracy by introducing personalized learning zones where \emph{learning devices} are grouped based 
on similar experiences as often observed in spatially near locations? 
More precisely, what impact does this have on heterogeneous 
and non-\ac{iid} data distributions?}
\end{tcolorbox}
\noindent To rigorously evaluate our approach under non-\ac{iid} conditions,
 we conducted extensive experiments comparing FBFL against baseline methods.
We systematically explored two distinct types of data skewness: 
 Dirichlet-based distribution (creating imbalanced class representations) and 
 hard partitioning (strictly segregating classes across regions).
The results reveal that baseline methods suffer from substantial performance degradation 
 under these challenging conditions.
These approaches consistently fail to capture the global objective and exhibit significant 
 instability during the learning process.
This limitation becomes particularly severe in scenarios with increased skewness, 
 where baseline models demonstrate poor generalization across heterogeneous data distributions.
\Cref{fig:exp-non-iid} presents key results from our comprehensive evaluation---which encompassed over $400$ distinct experimental configurations.
The first row shows results from the Extended MNIST dataset using
 hard partitioning across $6$ and $9$ distinct areas.
The performance gap is striking: baseline algorithms plateau at approximately $0.5$ accuracy, 
 while FBFL maintains robust performance above $0.95$.
Notably, increasing the number of areas adversely affects baseline models' performance, 
 leading to further accuracy deterioration.
In contrast, FBFL demonstrates remarkable stability, maintaining consistent performance 
 regardless of area count.
The second and third rows present results from Fashion MNIST and MNIST experiments, respectively.
While baseline methods show marginally better performance on these datasets 
 (attributable to their reduced complexity compared to EMNIST),
 they still significantly underperform compared to FBFL.
These comprehensive findings underscore the fundamental limitations of traditional approaches 
 in handling highly skewed non-IID scenarios.
The results provide compelling evidence for~\ref{itm:rq2}, demonstrating FBFL's superior capability 
 in maintaining high performance and stability across diverse data distributions through 
 its innovative field-based coordination mechanism.

\begin{figure*}
  \centering
  \begin{subfigure}{0.4\textwidth}
      \centering
      \includegraphics[width=\textwidth]{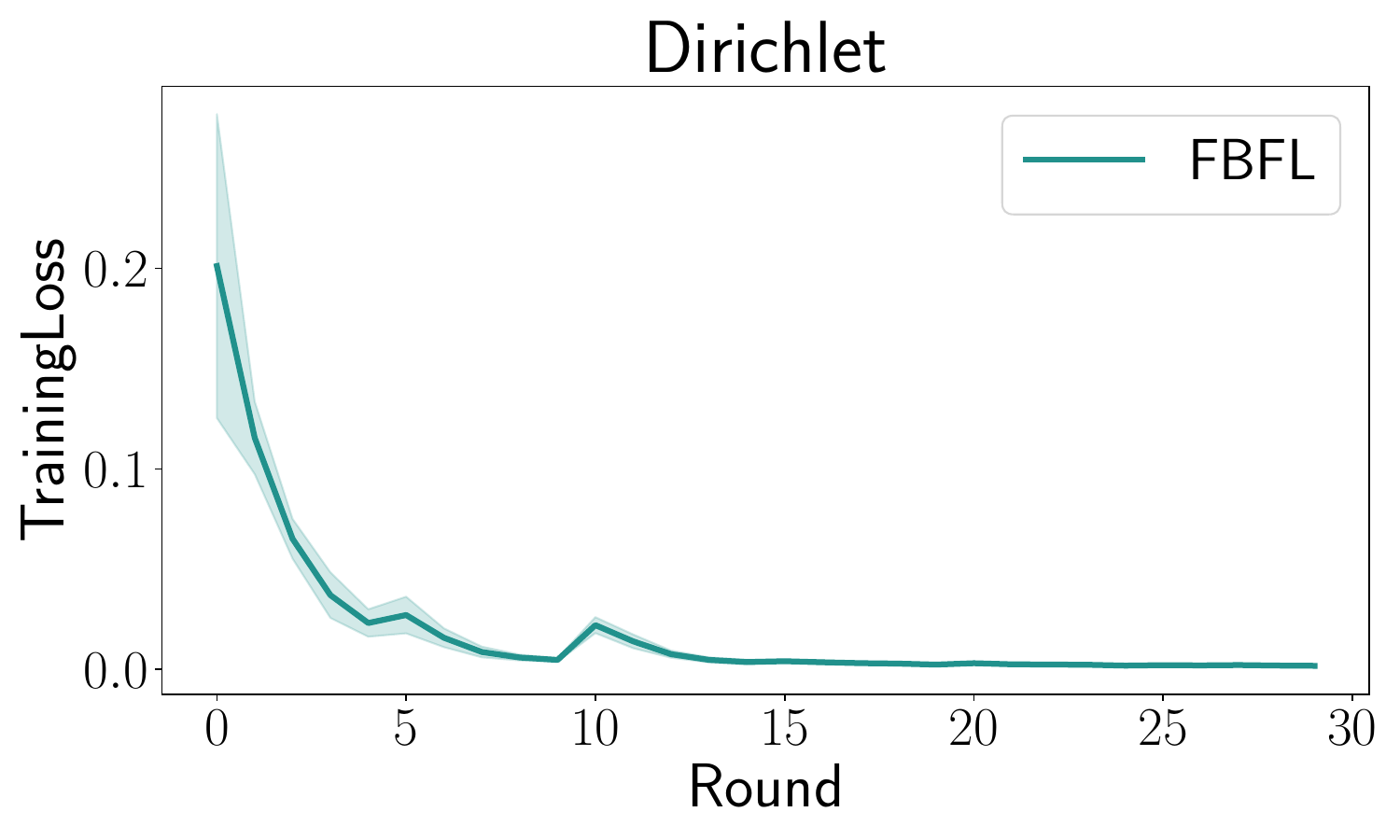}
  \end{subfigure}
  \begin{subfigure}{0.4\textwidth}
    \centering
    \includegraphics[width=\textwidth]{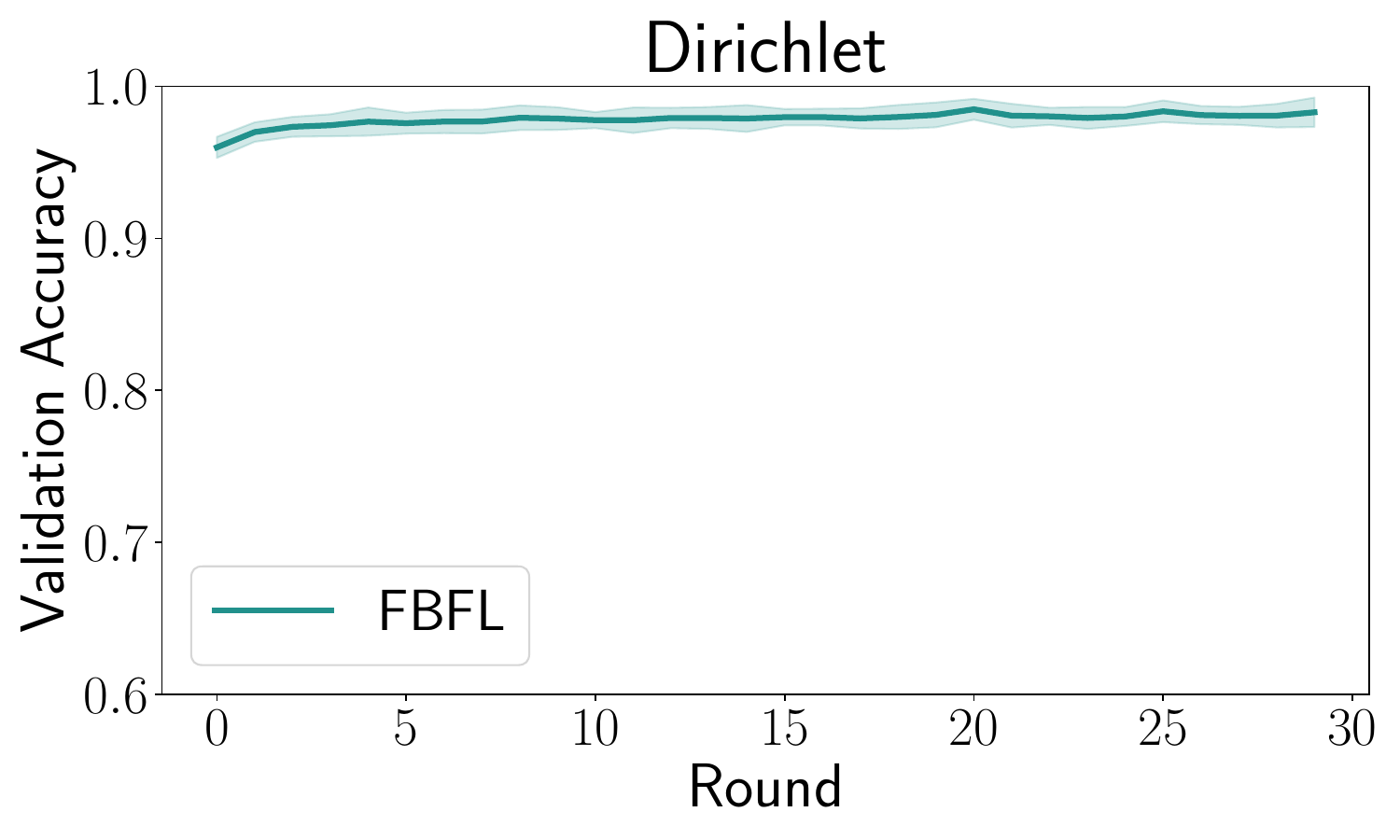}
  \end{subfigure}
  \caption{ Evaluation metrics in the scenario of aggregators failure. 
  It can be observed that, despite the failure of two aggregators, 
  the evaluation metrics do not undergo significant variations in either the training or validation phases, 
  and the learning process continues. 
  Notably, the only observable effect is a slight increase in the loss at training time step 10, 
  \ie{} when the aggregators fail and the system needs to re-stabilize,
  before it quickly resumes its downward trend. }
  \label{fig:exp-kill}
\end{figure*}

\begin{figure*}
  \centering
  \begin{subfigure}{0.32\textwidth}
      \centering
      \includegraphics[width=\textwidth]{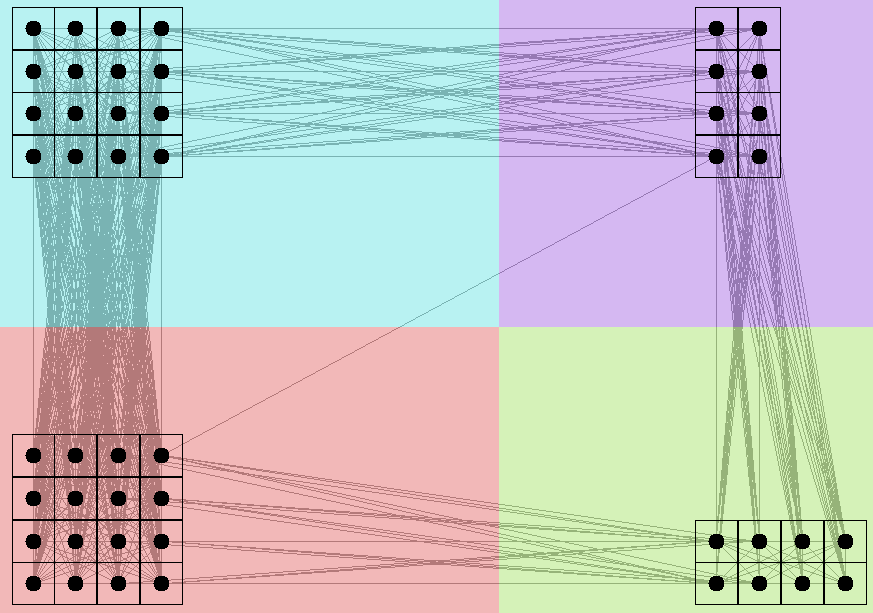}
      \caption{Start of the learning.}
      \label{fig:kill1}
  \end{subfigure}
  \begin{subfigure}{0.32\textwidth}
    \centering
    \includegraphics[width=\textwidth]{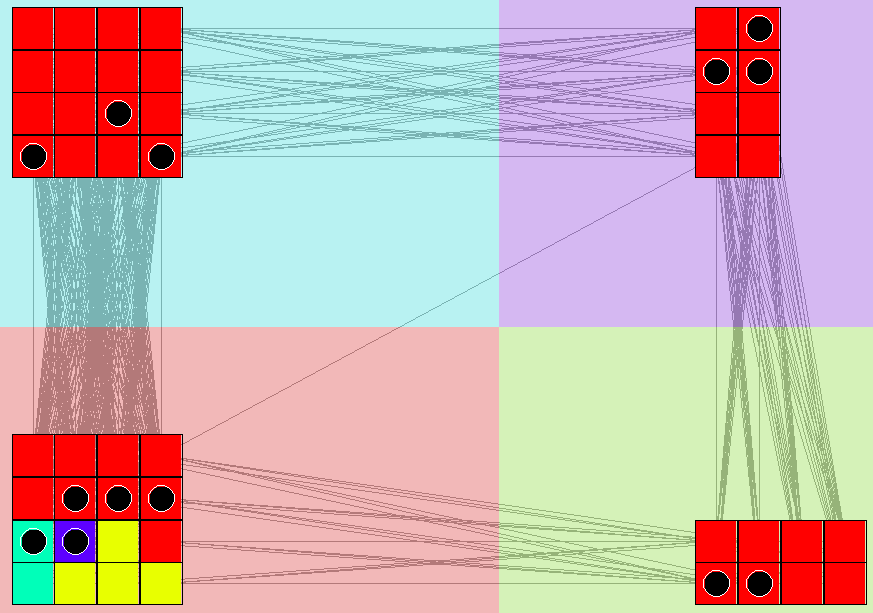}
    \caption{Subregions stabilization.}
    \label{fig:kill2}
  \end{subfigure}
  \begin{subfigure}{0.32\textwidth}
    \centering
    \includegraphics[width=\textwidth]{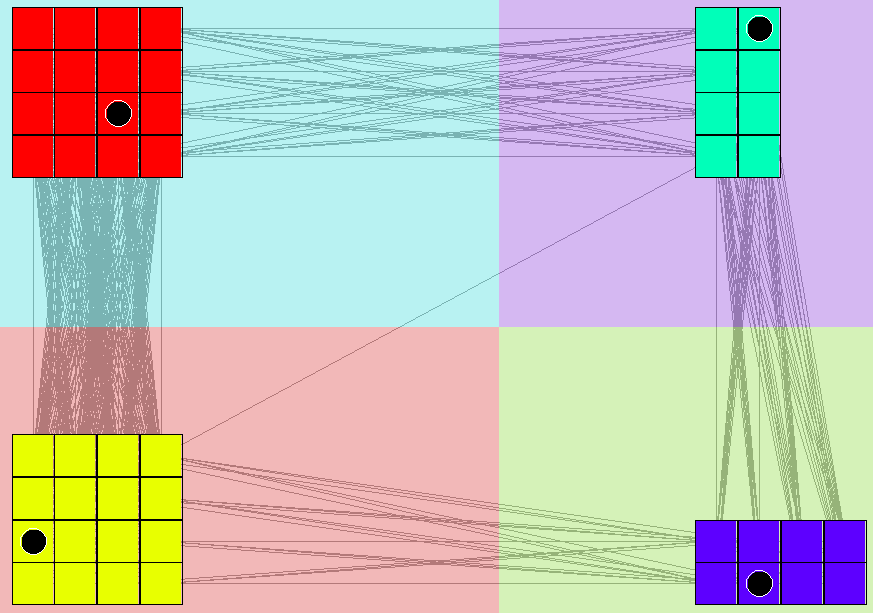}
    \caption{Learning.}
    \label{fig:kill3}
  \end{subfigure}
  \begin{subfigure}{0.32\textwidth}
    \centering
    \includegraphics[width=\textwidth]{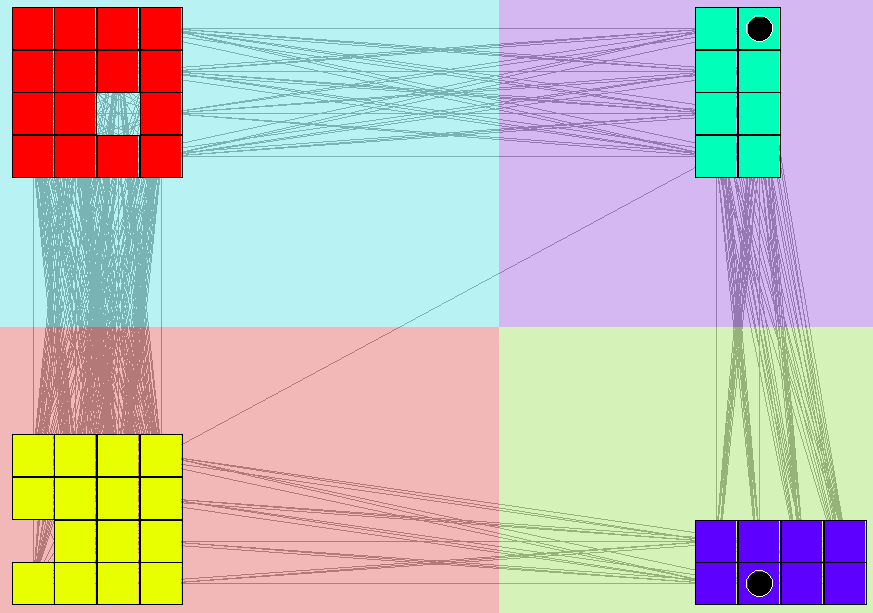}
    \caption{Aggregators failure.}
    \label{fig:kill4}
  \end{subfigure}
  \begin{subfigure}{0.32\textwidth}
    \centering
    \includegraphics[width=\textwidth]{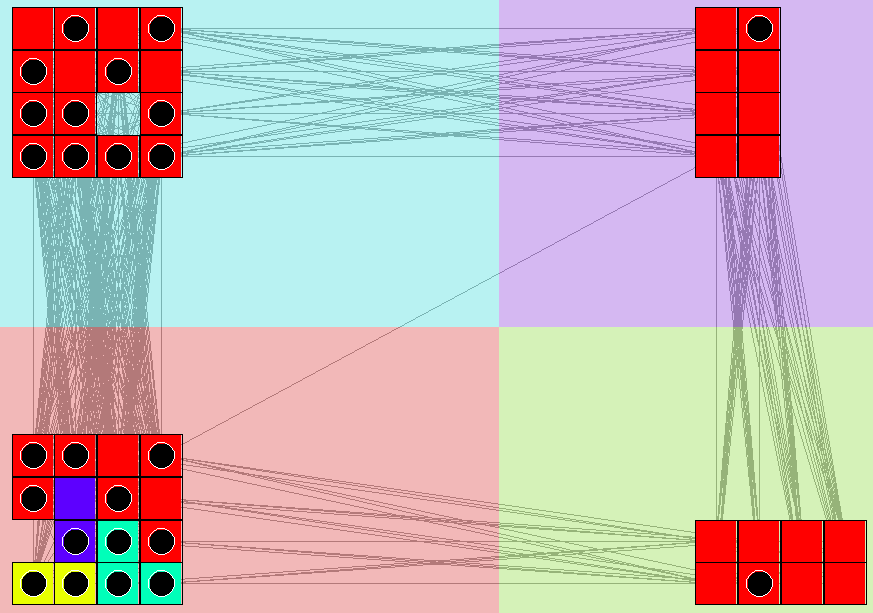}
    \caption{Subregions re-stabilization.}
    \label{fig:kill5}
  \end{subfigure}
  \begin{subfigure}{0.32\textwidth}
    \centering
    \includegraphics[width=\textwidth]{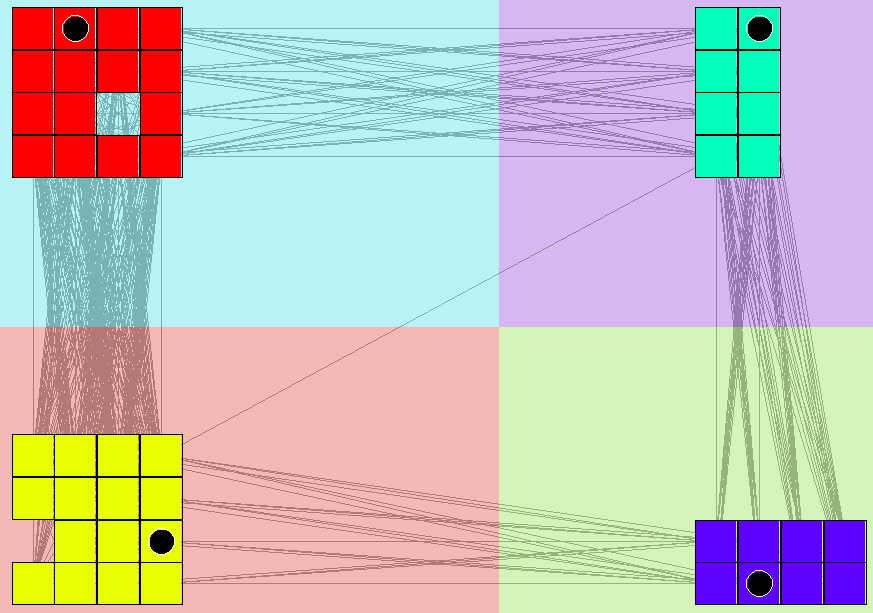}
    \caption{Learning.}
    \label{fig:kill6}
  \end{subfigure}
  \caption{ Graphical representation of the simulation from the Alchemist simulator.
  In this scenario, $50$ nodes (\ie{} the squares) are deployed in $4$ different subregions. 
  Background colors represent different data distribution, while nodes color represent 
  their respective federation.
  Black dots inside nodes represent aggregators.
  It can be observed that in unstable configurations, multiple aggregators 
  are present in each subregion, whereas, once the system stabilizes, only a single 
  aggregator remains per subregion.
  After the learning has started and the systems has found a stable configuration, we randomly 
  killed two aggregators node to simulate server failures.
  Notably, it is possible to see how the system is able to automatically re-stabilize. }
  \label{fig:kill-alchemist}
\end{figure*}

\begin{tcolorbox}[colback=gray!10!white, colframe=gray!80!black, title=\textbf{\ref{itm:rq3}}]
  \emph{What is the effect of creating a self-organizing hierarchical architecture through Field Coordination on Federated Learning in terms of resilience, adaptability and fault-tolerance?}
\end{tcolorbox}
\noindent The last experiment has been designed to evaluate the resilience of 
 the self-organizing hierarchical architecture proposed by FBFL (\ref{itm:rq3}). 
We simulated a scenario involving $4$ distinct areas and a total of $50$ devices. 
As for the other experiments, we ran $5$ simulations with varying seeds.
After allowing the system to stabilize under normal conditions (\Cref{fig:kill1,fig:kill2,fig:kill3}),
 we introduced a disruption by randomly killing two aggregator devices within the network (\Cref{fig:kill4}). 
The goal was to observe whether the system could autonomously recover, elect new aggregators, 
 and continue the learning process without significant degradation in performance.
The results, as depicted in performance charts (\Cref{fig:exp-kill}) and visualized in the 
 Alchemist simulator (\Cref{fig:kill-alchemist}), demonstrate that the FBFL architecture 
 successfully re-stabilizes into a new configuration. 
Specifically, the performance charts indicate only a minor increase in 
 loss during the killing phase (global round $10$), with no significant long-term drops in performance. 
Similarly, the Alchemist simulation shows a temporary transitional period following 
 the removal of the aggregators, during which federations are not correctly formed (\Cref{fig:kill5}). 
However, the system adapts by electing new aggregators, 
 and the configuration stabilizes once more (\Cref{fig:kill6}).

These findings highlight the robustness of the FBFL architecture, 
 emphasizing its capability to recover from disruptions and maintain stable performance. 
This resilience is a critical feature for real-world applications where system stability 
 under failure conditions is essential.

It can be observed that in unstable configurations, multiple aggregators are present in each subregion,
 whereas, once the system stabilizes, only a single aggregator remains per subregion.

\section{Limitations}\label{sec:limitations}

While the proposed approach demonstrates significant advantages in terms of adaptability and resilience, 
 it is important to acknowledge several limitations 
 that warrant consideration for future improvements.

\paragraph{\emph{Communication overhead and efficiency.}}
FBFL inherently requires more frequent communication exchanges compared to traditional centralized approaches, 
 since nodes must continuously exchange model updates with their neighbors:
 the gradient computation and collect-cast operations, while enabling self-organization, 
 introduce additional message overhead that scales with network density.
In resource-constrained environments or networks with limited bandwidth, 
this increased communication burden may hinder scalability and energy efficiency; nevertheless, 
FBFL generally incurs lower overhead than fully peer-to-peer schemes, where every client shares its model with all others. Future work will explore communication-efficient variants, 
  such as quantization techniques or sparse neural networks as discussed in recent works~\cite{domini2025sparse}.

\paragraph{\emph{Spatial-only clustering assumptions.}}
Our approach relies on the working hypothesis that spatial proximity serves as a proxy for data similarity.
This assumption holds in many IoT settings (\eg traffic management, air-quality monitoring) but does not universally apply.
In dense urban deployments, devices that are physically close may observe different phenomena (\eg{} indoor vs.\ outdoor sensors or distinct micro-environments), leading to suboptimal clustering.
Similar issues arise in hospitals, where co-located devices may monitor patients with heterogeneous conditions and clinical protocols.
The current implementation lacks mechanisms to detect and adapt to cases where 
 spatial clustering does not align with data distribution patterns.
Incorporating model-driven similarity metrics alongside spatial proximity could enhance 
the robustness of region formation, as discussed in subsequent works~\cite{DBLP:conf/acsos/DominiAFVE24} where 
the potential field and leader election metrics also consider model similarity 
based on cross-loss evaluation (without explicitly sharing data) leveraging a space-fluid approach~\cite{DBLP:journals/lmcs/CasadeiMPVZ23}.

\paragraph{\emph{Heterogeneous device capabilities.}}
Our current approach assumes relatively homogeneous computational capabilities across devices.
In practice, 
 federated learning environments often involve devices with vastly different 
 processing power, memory constraints, and battery life.
The dynamic leader election mechanism does not currently account for device capabilities, 
 potentially leading to resource-constrained devices being selected as aggregators, 
 which could degrade overall system performance.
However, future work could extend the leader election algorithm to consider device capabilities, 
 ensuring that aggregators are selected based on their ability to handle the computational load.

\section{Conclusions and Future Work}\label{sec:conclusions}

In this article, we presented \emph{Field-based Federated Learning (FBFL)}, leveraging field-based coordination to address key challenges of highly decentralized and reliable federated learning systems.
Our approach innovates by exploiting spatial proximity to handle non-IID data distributions and 
 by implementing a self-organizing hierarchical architecture through dynamic leader election.
FBFL demonstrates robust performance in both IID and non-IID scenarios while providing inherent resilience against node failures.
Our results show that forming personalized learning subregions mitigates the effects of data skewness.
Compared to state-of-the-art methods like Scaffold and FedProx, FBFL significantly outperforms in 
 scenarios with heterogeneous data distributions.

Future work could explore advanced field coordination concepts, such as space fluidity~\cite{DBLP:journals/lmcs/CasadeiMPVZ23}, 
 to enable dynamic segmentation of learning zones based on evolving trends in the phenomena 
 being modeled, rather than relying on static, predefined assumptions
Additionally, the potential for FBFL to support personalized machine learning models in 
 decentralized environments offers promising applications across domains such as 
 edge computing and IoT ecosystems.
Finally, another avenue for improvement lies in integrating sparse neural networks to reduce resource consumption, 
 making the approach more efficient and scalable for resource-constrained devices.

\section*{Acknowledgment}
Lukas Esterle was supported by the Independent Research Fund Denmark through the FLOCKD project under the grant number 1032-00179B.
Gianluca Aguzzi was supported by the Italian PRIN project ``CommonWears'' (2020 HCWWLP).
Finally, this work was also supported by FAIR-Future Artificial Intelligence Research” project 
 (PNRR, M4C2, Investimento 1.3, Partenariato Esteso PE00000013), funded by the European 
 Commission under the NextGenerationEU programme.
\bibliographystyle{alphaurl}
\bibliography{bibliography}

\end{document}